\documentclass[9pt,compsoc,peerreview]{IEEEtran}
\usepackage[sort,nospace]{cite}
\usepackage{graphicx}
\usepackage{amsmath}
\usepackage{amssymb}
\usepackage[ruled,vlined]{algorithm2e}
\newcommand{\bi}{{\bf i}}
\newcommand{\bs}{{\bf s}}
\newcommand{\bc}{{\bf c}}
\newcommand{\bk}{{\bf k}}
\newcommand{\bE}{{\bf E}}
\newcommand{\bW}{{\bf W}}
\newcommand{\bS}{{\bf S}}
\begin{document}

\title{Capturing spatial interdependence in image features: the counting grid, an epitomic representation for bags of features}



\author{Alessandro~Perina,
        Nebojsa~Jojic \\
Microsoft Research Redmond, USA
\IEEEcompsocitemizethanks{\IEEEcompsocthanksitem A. Perina and N.Jojic are with Microsoft Research, Redmond, WA.\protect\\
E-mail: alperina@microsoft.com}% <-this % stops a space
\thanks{}}

\IEEEcompsoctitleabstractindextext{
\begin{abstract}
In recent scene recognition research images or large image regions are often represented as disorganized ``bags'' of features which can then be analyzed using models originally developed to capture co-variation of word counts in text. However, image feature counts are likely to be constrained in different ways than word counts in text. For example, as a camera pans upwards from a building entrance over its first few floors and then further up into the sky Fig. \ref{fig:figure0}, some feature counts in the image drop while others rise -- only to drop again giving way to features found more often at higher elevations. The space of all possible feature count combinations is constrained both by the properties of the larger scene and the size and the location of the window into it. To capture such variation, in this paper we propose the use of the counting grid model. This generative model is based on a grid of feature counts, considerably larger than any of the modeled images, and considerably smaller than the real estate needed to tile the images next to each other tightly. Each modeled image is assumed to have a representative window in the grid in which the feature counts mimic the feature distribution in the image. We provide a learning procedure that jointly maps all images in the training set to the counting grid and estimates the appropriate local counts in it. Experimentally, we demonstrate that the resulting representation captures the space of feature count combinations more accurately than the traditional models, not only when the input images come from a panning camera, but even when modeling images of different scenes from the same category.
\end{abstract}
\keywords{ BBag of Features \and Spatial Layout \and Scene Analysis}
}

\maketitle

\IEEEdisplaynotcompsoctitleabstractindextext

\IEEEpeerreviewmaketitle

\begin{figure}[t!]
\centering
\includegraphics[width=0.8\columnwidth]{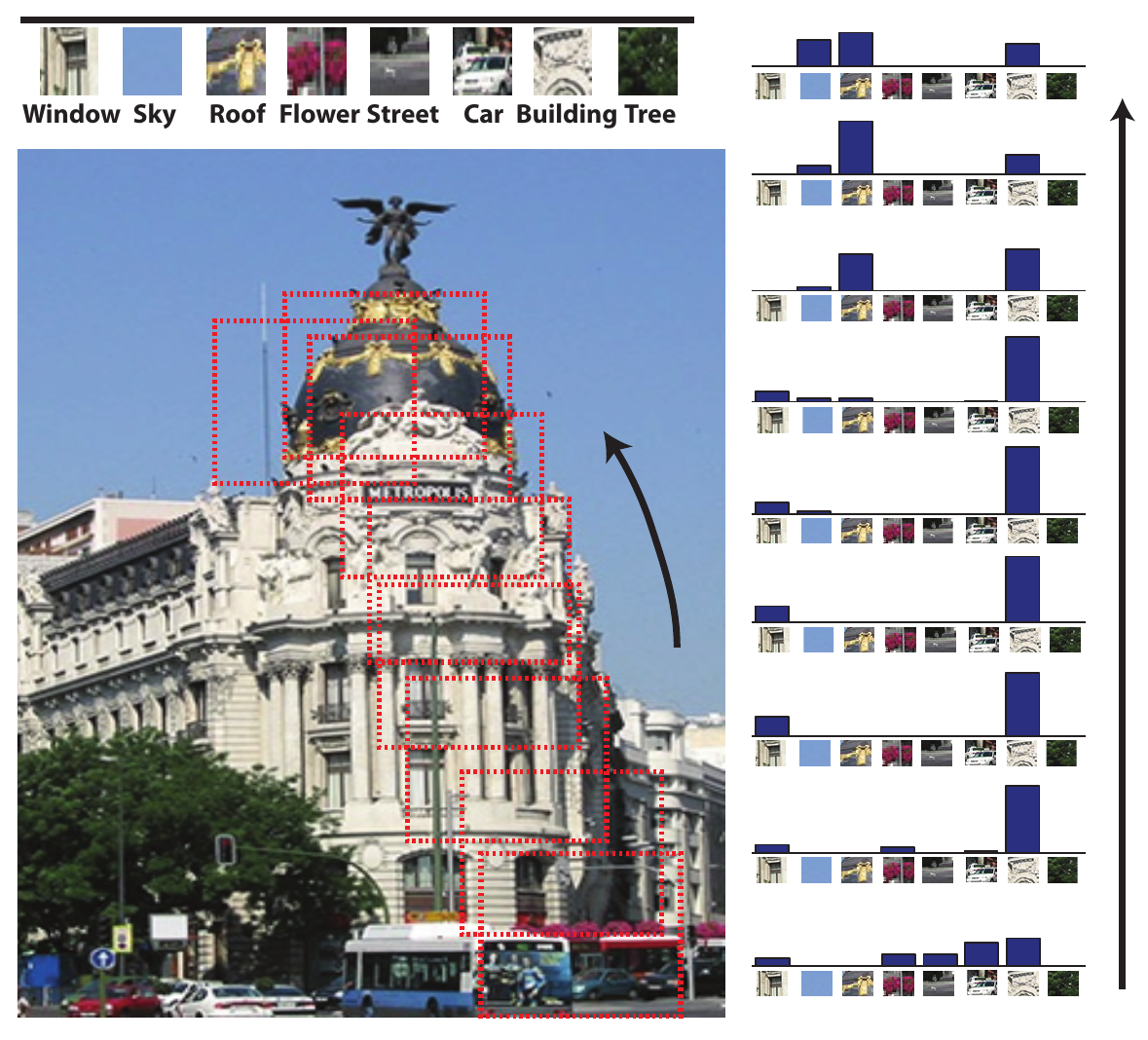}
\caption{Feature counts change slightly as the field of view moves. For example, the abundance of the ``car'' features is reduced,  but the counts of the features found on building facades are increased. The counting grid model accounts for such changes naturally, and it can also account for images of different scenes.}
\label{fig:figure0}
\end{figure}

\section{Introduction}

\begin{figure*}[t]
\centering
\includegraphics[width=1\textwidth]{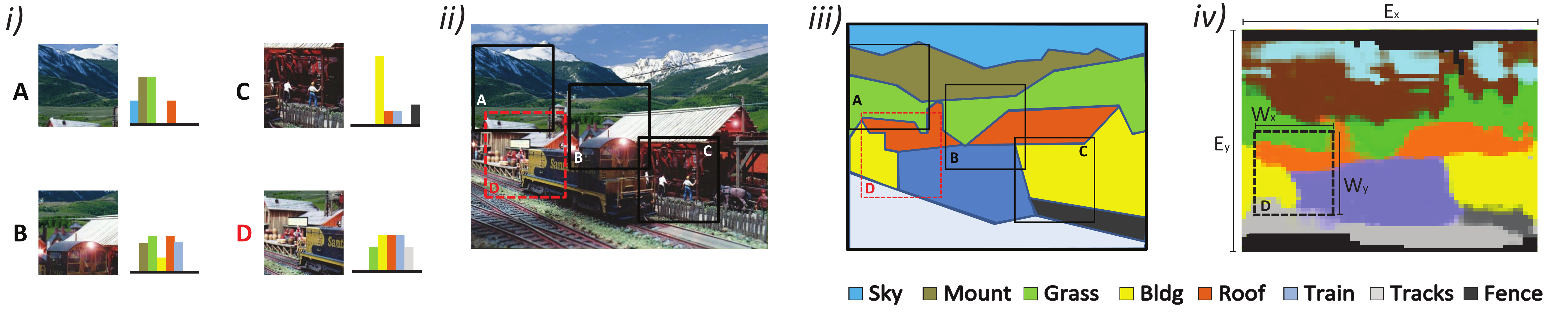}
\caption{Counting grid illustration. i) Images and their Bag of feature representation. ii) Images of a train station, taken as windows into the larger scene. iii) Hand labeled features. iv) Scene reconstructed (e.g., counting grid) starting from bags taken from 50 windows (see the Text for details).}
\label{fig:figure1}
\end{figure*}

A popular way to deal with diversity of imaging conditions as well as geometric variation in objects or entire scenes is to simply represent images or image regions as disordered ``bags'' of image features \cite{bow,bowReview,ldaVis}. These models are particularly, attractive due to the computational efficiency and simplicity achieved by ignoring spatial relationships of the image patches or object parts. 

The bag of features can arise in a variety of ways. For example, after extracting local low-level features from images, these are often clustered and a discrete ``codeword'' is assigned to each feature descriptor. An image is then described by a histogram over the codebook entries. Ideally, these features should be highly discriminative so that most categories of images of interest are uniquely identifiable by the presence of a handful of features. In practice, however, individual features are not sufficiently discriminative, and modeling joint variation in feature counts becomes an interesting machine learning problem. 

It is tempting to use here the existing discrete models, such as histograms \cite{bowText}, multinomial mixtures \cite{5640674,mixtureUnigrams} or topic models \cite{lda,pLSA}, already extensively validated on text data, where each document is also simply represented as a count distribution over the entire vocabulary. However, the bags of features extracted from natural images have an imprint of the images' spatial structure, which is evident when the bags from related images are considered \emph{together}. Thus ignoring these natural constraints on the feature counts may have negative consequences in classification tasks.

For an illustration, Fig. \ref{fig:figure1} provides a synthetic example starting with several images of a train station, taken as windows into the larger scene - \textit{ii)}. Just for illustrative purposes, we hand-labeled the scene with feature labels as shown in - \textit{iii)}. In a realistic application, where we may want to train a model that assigns high likelihood to images of train stations, it is likely that most available images would be taken with a narrower field of view, as simulated here.  Feature extractors would presumably generalize much less effectively than our ideal features, but still enough to permit comparisons of images of \emph{different} train stations, too. Then the question is if a learning model that captures feature count co-variation uses the training data efficiently. Assuming that a few images are taken at random from the scene, we wonder if the feature counts in these images are sufficient to predict the possible feature counts in other images of the scene. In particular, we consider images taken from the regions close to A, B, and C and ask the question if the image D would fit the so defined train station class. 

The literature uses two sets of approaches to this problem. Kernel or nearest-neighbor techniques start with the comparisons of the feature counts in the test image and each of the previously studied exemplars \cite{citeulike:2323643,spk,defenseKnn,vs,SpatialEnvelope}. Although this comparison can be done in many different ways, we note here that these approaches would be complicated by the fact that none of images A,B and C have the combination of all five features that are present in D (see Fig. \ref{fig:figure1}-\emph{i)}). The other approach is to consider all bags of features together and generalize \cite{ldaVis,scPlsa,4032602,5640674,bow,bowRec,torralbaMoG}. A simplest approach to this would be to simply merge the bags. In this case, there is a danger of overgeneralization. For this particular example, there is a need for interpolating between the feature count vectors for A,B,C. However, this interpolation is best performed by spatial reasoning. Across various windows into the scene we find that from the top of the window to the bottom we sometimes see roof, train, tracks, in that order, but other times we see mountain, grass, roof, train. We can infer that the grass, roof, train, tracks combination is likelier than the existence of the mountain, roof, train, tracks combination of features. Furthermore, the proportions of different features in the images carry information about the thickness of the layers of these features, which should be useful for inferring which previously unseen feature count combinations can be found elsewhere in the scene. We show in this paper that, surprisingly, not much of the spatial organization of the features in the training images needs to be retained in order to perform the spatial reasoning about which feature combinations are likely.

In Fig. \ref{fig:figure1}-\textit{iv)} we show the counting grid inferred by iterating Eqs. \ref{eq:qkl2} and \ref{eq:M2} on the label counts from 50 windows into the scene taken at random, \emph{but avoiding all windows that contain all five of the features in D in any proportion}. Each training image was represented as a set of $2\times 2$ feature bags (upper left, lower left, upper right, lower right). Without using the original window location information, the counting grid was computed so that for each training image, a window into the counting grid can be found so that the appropriate sections have matching histograms. The resolution of the reconstructed feature layout of the large scene goes well beyond what would be expected from a crude $2 \times 2$ tessellation of the input images (the height of each section is roughly 20\% of the large scene and only the feature counts in each section were used, not their spatial layout within the section). Although none of the training examples was taken from the area close to D where all five of D's features can be seen in a single image, that part of the scene is reconstructed as well, and D's histogram can be matched well. \\
\begin{figure*}[t]
\centering
\includegraphics[width=1\textwidth]{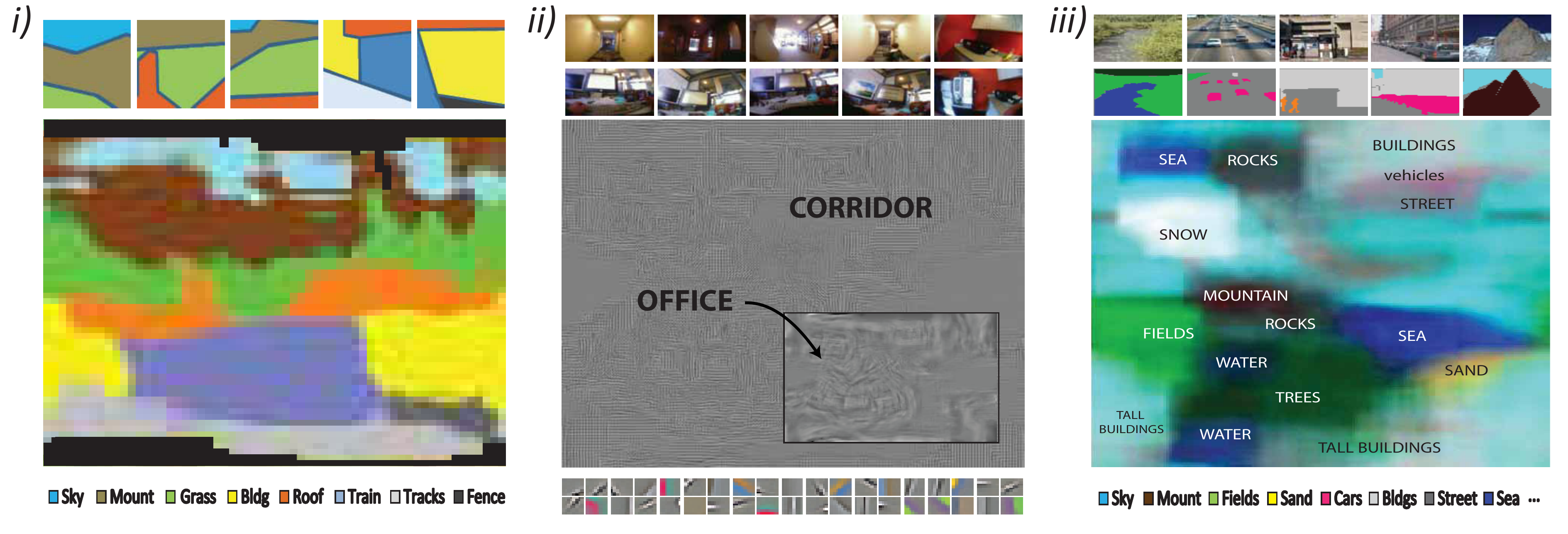}
\caption{Illustration of counting grids using different data at different levels of abstraction. At the top of each CG we show few examples of the training images from which we extracted the bags. The bottom row of each panel, illustrates the features. i) A counting grid learned using patches extracted from the train station toy example of Fig. \ref{fig:figure1}. In this case all the bags come from windows into the same image, which is reconstructed on the grid. ii) A counting grid estimated from images taken with a wearable camera \cite{slcg}. In this case we learned a dictionary of features (illustrated by the textons on the bottom) clustering image patches, as in \cite{coates}. To illustrate the office scene (e.g., the computer screens - see images on the top), we overlap these textons by as much as the patches were overlapping during feature extraction process, and then average to create a clearer visual representation. iii) A counting grid estimated starting from LabelMe annotations (see the top row) \cite{labelme}.}
\label{fig:figure2}
\end{figure*}

In this simple example, the training images are different views of a single scene. However, at the feature level, images of other train stations are likely have a similar layout, and so they could be used to learn a counting grid. 
In practice, we rarely have access to highly discriminative and reliable features, and so instead of the 8 fake features in our example, in our experiments we had to use hundreds of simpler automatically derived  features, and infer the counting grids from related images of different scenes. For example in Fig. \ref{fig:figure2}-\emph{iii)} we used as features the human-supplied labels for LabelMe \cite{labelme} dataset, and in - \emph{ii)} the outputs of hundreds of simple computational feature detectors applied to images from various scenes in the SenseCam dataset \cite{slcg}. As opposed to the train station example, input images are not subimages of a larger single scene, but rather images of the same types of scenes.
Each window into a counting grid represents a possible feature combination\footnote{The CG model also learns a prior over the grid window usage which may prevent some combinations.} present in the dataset (\ref{fig:figure2}- \emph{ii)}) and the model is able to reconstruct the feature layout only exploiting the spatial patterns very coarsely, but  depending on feature count co-variation for most of its reasoning power. For example, despite only relying on a 2$\times$2 tessellation of each input image, full resolution panoramas of office and corridors are visible in the CG in Fig. \ref{fig:figure2}-\emph{ii)}.

This paper presents and extends the counting grid model \cite{Jojic11,cgCvpr}. The basic model is extended to include priors which help with overfitting issues. We also formally introduce the tessellated counting grid model and analyzed the the extreme tessellations where each bag captured a single feature, collapsing the representation into the discrete epitome \cite{epitomeLr}.  The paper also provides full comparisons of different algorithms on various datasets, including the effect of grid and window size variation can be found in the experimental section.

Specialization and extensions of the counting grid model already appeared in top tier conferences \cite{slcg,ccg,DBLP:conf/cvpr/AmerT12,cgFlickr}. Nevertheless, in this paper we want to limit our attention to the basic variants of the counting grid model, their properties and relationships with the standard techniques for modeling bags of words in computer vision \cite{ldaVis,bow,bowRec}. We found that our representation captured the space of possible feature count combinations for various image categories significantly better than other generalization techniques, and that our simple generative model, which can be used for unsupervised learning and clustering, too, often rivals the state of the art based on discriminative techniques that require supervision.

\begin{table*}[t!]
 \caption{Generative approaches to scene analysis. }
 \label{tab:tabintro}
  \begin{center}
\begin{tabular}{l|ccc|ccc}
\emph{ Method} & \multicolumn{3}{c}{\emph{Componential hidden variables}} &  \multicolumn{3}{c}{\emph{Spatial structure in input}}  \\
  & Single integer & Multiple integers &  Continuous & BoW & Tessellated & Pixel \\
  \hline
  LDA \cite{lda,ldaVis} & & &  \checkmark & \checkmark &  (\checkmark)  & \\
  Multinomial Mixtures \cite{torralbaMoG,bow} & \checkmark &  &  & \checkmark &  & \\
  Spatial BoW \cite{bowReview} & \checkmark &  &  & & \checkmark  & \\
  Reconfigurable BoW \cite{bowRec} & & \checkmark &  & & \checkmark  & \\
  Epitomes \cite{epitome,epitomeLr,epitomeSca} & \checkmark & &   &  &  & \checkmark \\
  Flex. Sprites \cite{flex} &  & \checkmark &   &  &  & \checkmark \\
  \textbf{Counting grids} & \checkmark & & &  \checkmark &  &  \\
  \textbf{Tessellated Counting grids} & \checkmark & & &   & \checkmark  &  \\
  \hline
\end{tabular}
  \end{center}
\end{table*}
\subsection*{Related Work}
% A: Aggiungi un po' di discriminativi...

Previous probabilistic approaches to scene recognition treat the spatial arrangement of image features in different ways. 

In bag of words (BOW) models \cite{bow}, spatial relationships among features are completely ignored in order to facilitate computational efficiency and high level of generalization. 
Topic models \cite{lda,ldaVis,pLSA}, for example, assign a topic to each codeword based on their co-occurrence and describe images as admixtures of topics. Another bag of word model is described in \cite{torralbaMoG}, where a scene model is a mixture of Gaussians model trained on the gist descriptors \cite{gist}. As we will see in this paper the basic counting grid model \cite{cgCvpr}, also reduces to a (large) mixture, but with highly tied parameters, reflecting the inherent spatial structure of the data. Each bag is represented as a point in a large grid of feature counts. This latent point is a corner of a window of grid points which are uniformly combined to match the (normalized) feature counts in the image.

To capture some spatial information, it is possible to separate the bags originating in different (pre-defined or learned \cite{randomizedSpatialPartition,DBLP:journals/ivc/PerinaCM10,citeulike:9350037}) regions of the image. These models are sometimes referred to as spatial-BoW models \cite{bowReview,bowRec}. The tessellated counting grids that we introduce here also have that flavor, although in our approach tessellation helps guide the quilting of the bags of words reconstructing the layout with sub-region accuracy.  Thus tessellated counting grids capture layout-driven constraints on counts within the regions, even though this information is not directly provided during learning: The layout within a region of one image is inferred based on the feature distributions found in regions of many other images, assuming that misalignments of these images are often smaller than the size of the tessellated regions. In contrast, typical spatial-BOW models requires the modeled images to be approximately aligned. Recent approaches that relax this assumption are \cite{bowRec,dpmScenes,slcg}. The former, the Reconfigurable BoW model \cite{bowRec}, represents a scene as a collection of parts arranged in a reconfigurable pattern. Each image is divided into pre-defined regions and a latent variable specifies which ``region model'' (e.g., sky, grass...) is assigned to each image region. On the other hand, \cite{dpmScenes,slcg} represents scenes using deformable parts. In \cite{dpmScenes} a lower-resolution root filter is placed in the center of the image and a set of higher-resolution part filters arranged in a flexible spatial configuration.
% The Spring Lattice model presented in \cite{slcg}, is an extension of the tessellated counting grid model. It maps each sectors to different sub-windows in the counting grid in a configuration that is close, but not exactly the same as the configuration of the source sectors. \\

It is also possible to keep the spatial arrangement of features intact, sacrificing some generalization in the basic representation of the input, and allowing the model to capture the problems with this rigidity through various levels of uncertainty modeling.  For example, the epitome-like models \cite{epitome,epitomeLr,epitomeSca} quilts images or image patches, essentially building giant panoramas consisting of probability distributions in each location. As these are based on pixel-to-pixel comparisons they cannot generalize well in case of large geometric deformations, and so they are mostly used to model relatively small image patches, typically for synthesis, or modeling large scenes or textures that can tolerate the lack of transformation invariance beyond translation \cite{epitomeSpatialized,flex,prlEpitome}. Epitomes have been employed in scene analysis, only on particular datasets where ``panoramic stitching'' would work, e.g., sequences taken with wearable cameras \cite{epitomeLr,epitomeSca}.

Various scene modeling techniques also take different approaches to representing componential structure of natural scenes. Being ad-mixtures, rather than simple mixtures, topic models \cite{lda,pLSA,ldaVis} are a simple example of multi \emph{-part} or \emph{-object} models. Other examples of componential models, are the flexible sprites model \cite{flex}, which allow each image to be mapped to multiple sources and \cite{bowRec} in which each sector is mapped independently.  On the other hand, the counting grids, epitomes \cite{epitome,epitomeLr,epitomeSca}, histogram-based approaches \cite{torralbaMoG,bow} are essentially mixtures because they map the entire scene to a single point: a position or a mixture component. (These models can, however, be turned into ad-mixtures.)

The main topic this paper is modeling bags of features in computer vision. In the experimental section we will mainly consider generative approaches and compare counting grids with latent Dirichlet allocation \cite{ldaVis}, mixture models \cite{torralbaMoG}, epitomes \cite{epitomeLr} and the reconfigurable bag of words model \cite{bowRec}. The nature of each generative approach just discussed is summarized in Tab. \ref{tab:tabintro}. It should be noted however that the models presented here can be used as components in hierarchical models, and that the basic idea of modeling intersections and laying them out on an inferred grid can be used within other machine learning techniques, including non-generative approaches.

\section{Imprint of spatial organization in disordered bags of words}

As discussed above, we would like to understand the hidden constraints that govern the often-practiced simplification of images into bags of features. This simplification has two stages. First, image features $z_{i,j}$ are extracted on a grid inside the image. These features are discrete, $z \in [1..Z]$, and they point to a codebook of features obtained by clustering the multidimensional real-valued features calculated by local image processing, e.g., SIFT \cite{sift}. Next, the feature counts are computed $c_z=\sum_{\bi} [z_{\bi}=z]$, where $[\cdot]$ is the indicator function. Only the counts $c_z$ are then retained, and the spatial distribution $z_{\bi}$ is typically forgotten, with the justification that establishing correspondence for individual image locations across different images of the same thing would be prohibitively expensive, and that in practice only the presence or absence of features is informative, not their spatial distribution.
However, if we consider a set of such bags of words from related images we can see that the feature counts in these disordered bags of features may still indirectly follow the rules of spatial organization. For example, if the bags $\{c_z^t\}$, indexed by $t$ are extracted from several overlapping windows from a larger image, then the spatial structure of that image is imprinted in the particular count combinations in these bags. Furthermore, the spatial layout of the features in the large image may even be recoverable from these disordered bags! If the bags $\{c^t_z\}$ are created from \emph{all} the overlapping windows from a large image, and if the source location for each bag is known, then we can easily see that under minimal additional assumptions regarding the boundaries in the image, we can reconstruct feature indices $z$ at each location in the large image by solving the system of linear equations that arise from the count constraints. Consider two horizontally neighboring windows: The count differences are completely determined by the feature identities of the only two columns that the two do not share. To separate the effect of the two columns, we can consider another pair of overlapping images whose count differences depend on only one of those two columns. To further break each column apart, we can consider vertically neighboring windows, etc. As long as the image has a thick enough border with only a single feature present, we can propagate these constraints until any given location's feature is uniquely determined.\\
 In this way, we can reconstruct a large grid of features such that any of the count combinations we see in the given bags can be found in an appropriate window in this reconstruction. But this implies that the bags of features from the images of the same scene, when considered jointly, obey very strong constraints and thus taking these constraints into account will likely improve image analysis tasks that depend on the feature count representations. This insight leads to several interesting problems which we address in the next section.
\begin{itemize}
\item \emph{Joint estimation of the feature layout and the matching of the bags to windows into it}: If the bags of features (feature counts) from many -- but \emph{not all} -- overlapping windows from a large scene are provided, and if the original locations of these windows are \emph{withheld}, can we still reconstruct at least some of the original spatial arrangement of the features? \\
\item \emph{Category modeling}: If the bags of features are not coming from the windows into a single scene, but instead from different but related images (e.g. of a particular image category or an object class), would these bags, when considered jointly, imply some spatial layout of the features, and would this layout help predict which combinations of feature counts are more likely in bags of features extracted from new images of the category in question? \\
\item \emph{Using more of the original structure}: Given that in practice we typically have access to the original images, can more of their spatial structure be used in learning the spatial layout of features that would in turn constrain the bag of words representation in a useful way? \\
\end{itemize}

\section{The counting grid model}

\begin{figure}[t]
\centering
\includegraphics[width=0.8\columnwidth]{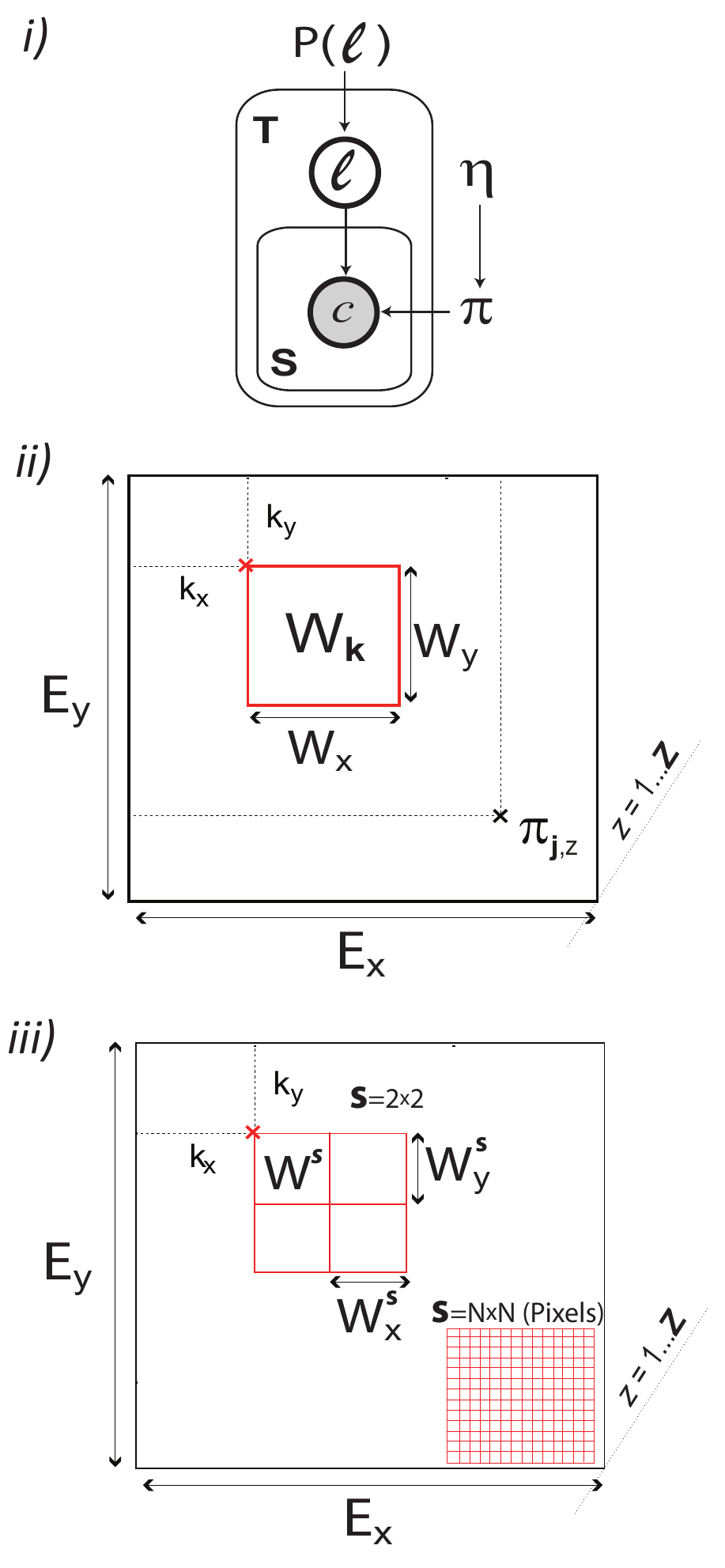}
\caption{i) Counting grid Generative model, ii) Counting grid Geometry, iii) Tessellated counting grid geometry.}
\label{fig:gm}
\end{figure}

The basic counting grid $\pi_{\bi,z}$ is a set of normalized counts of features indexed by $z$ on the grid $\bi = (i_x,i_y) \in \bE = [1\dots E_x] \times [1\dots E_y]$, with $\sum_z \pi_{\bi,z}=1$ everywhere on the grid \cite{Jojic11,cgCvpr}. 
A given bag of image features, represented by counts $\{c_z\}$ is assumed to follow a distribution found somewhere in the counting grid. In other words, the bag can be generated by firstly averaging all counts in the window $W_\bk$ of size $W_x \times W_y$ placed at location $\bk$
\begin{equation*}
W_\bk = [k_x, \dots, k_x+W_x-1] \times [k_y, \dots, k_y+W_y-1]
\end{equation*}
to form the histogram
\begin{equation} \label{eq:h}
h_{\bk,z}=\frac{1}{\left( W_x\cdot W_y\right)}\cdot \sum_{\bi \in W_{\bk}} \pi_{\bi,z}
\end{equation}
and then generating the features in the bag. The sum in Eq. \ref{eq:h} is carried out in all the locations $\bi$ in the window $W_\bk$. 
An example of counting grid geometry is illustrated in Fig. \ref{fig:gm}-\emph{ii)}
In other words, the position of the window in the grid is a latent variable $\ell$ given which the probability of the bag of features $\bc = \{c_z\}_{z=1}^Z$ is
\begin{equation}
p( \bc | \ell = \bk )= \prod_{z=1}^Z \big( h_{\bk,z}\big)^{c_z} = \alpha \cdot \prod_{z=1}^Z \big( \sum_{\bi \in W_{\bk}} \pi_{\bi,z}\big)^{c_z}
\end{equation}
where the constant $\alpha= (\frac{1}{W_x\cdot W_y})^{\sum_z c_z}$ .
 In our notation the letter $\ell$ indicates the latent variable, while $\bi$ and $\bk$  a generic position in the grid. 

The Bayesian network of the model is illustrated in Fig. \ref{fig:gm}-\emph{i)}. For a given grid $pi$, it defines the following joint distribution over all bags of features $\{c_z^t\}$, indexed by $t$ and their corresponding latent window positions $\ell^t$ in the counting grid
\begin{equation}
P\big( \{\bc^t\},\{ \ell^t \}  \big) \propto   \prod_{t=1}^T \sum_{\bk \in \bE } \left( P(\ell^t = \bk)\cdot \prod_{z=1}^Z \big( \sum_{\bi \in W_{\bk}} \pi_{\bi,z}\big)^{c^t_z} \right) \nonumber
\end{equation}
Where $P(\ell = \bk)$ represents the overall prior probability of a mapping location. The first sum in the RHS is performed over all the location of the counting grid, while the second over all the locations in the window placed at location $W_\bk$.

To summarize the notation we will use throughout the paper, $\ell$ is the hidden variable that represents the mapping location in the grid; each bag (sample) is mapped to a (possibly) different location and we will use the superscript $t$ to refer to the particular $t$-th bag therefore $\ell^t$ will represent the mapping position of the bag $\bc^t=\{c_z^t\}$. Since we are introducing a probabilistic model, it is interesting to estimate the prior probability $p( \ell^t = \bk )$ of mapping the $t$-th sample to location $\bk$. In this case $\ell^t$ is the $t$-th sample's hidden variable (window location), while $\bk$ is a generic constant or index that represents a possible location in the grid that  samples share. This distribution is a table over all values for $\bk$ and shared across all samples (independent of $t$). On the other hand, the posterior distribution $p(\ell^t=\bk | \bc^t)$, or its (exact) variational counterpart $q(\ell^t=\bk)$, is a function of the counts seen in the $t$-th sample and capture the quality of the fit to different windows in the grid of the $t$-th sample in particular. 

\subsection{Inference and learning}

To compute the log likelihood of the data, $\log P$, we need to sum over the latent variable $\ell$ before computing the logarithm, which, as in mixture models, or as in epitomes \cite{epitome}, makes it difficult to perform assignment of the latent variables while also estimating the model parameters. Although the following is the exact EM procedure, we use the variational \cite{variationalJordan,variational} notation $p(\ell^t | \bc^t) = q( \ell^t)$, and bound (variationally) $\log P$ (omitting the effect of additive constant that arises from $\alpha$)  to derive an iterative EM algorithm:
\begin{eqnarray}
\log P & \ge & \sum_{t=1}^T \sum_{\bk \in \bE} \Bigg( q( \ell^t = \bk ) \cdot \log q( \ell^t = \bk ) \phantom{\Bigg)} \nonumber \\
&-& \phantom{\Bigg( } q( \ell^t = \bk)\cdot\log P(\ell= \bk) \phantom{\Bigg)} \nonumber \\
&-& q( \ell^t = \bk ) \cdot\Big( \sum_z c^t_z \cdot\log h_{\bk,z} \Big) \Bigg) = B, \label{eq:bound}
\end{eqnarray}
Because of the use of fully parameterized $q$, optimizing the bound is equivalent to optimizing the log likelihood of the data, as long as the $q(\ell^t)$ distributions are also optimized. 
Keeping the model parameters fixed, optimizing these $q$ distribution (exact E step) leads to
\begin{equation}
q(\ell^t = \bk )\propto P(\ell = \bk)\cdot \exp \Big( \; \sum_{z=1}^Z c^t_z \cdot \log h_{\bk,z} \Big), \label{eq:qkl}
\end{equation}
which simply establishes that the choice of $\ell$ should minimize the KL divergence between the counts in the bag and the counts $h_{\bk,z}$ in the appropriate window $W_{\bk}$ in the counting grid. For each $t$, the above expression is normalized over all possible window choices $\bk$. 

To optimize the bound $B$ with respect to model parameters (M step) we note that the first term in Eq. \ref{eq:bound} involves these parameters, and it requires another summation before applying the logarithm. The summation is over the grid positions $\bi$ within the window $W_\bk$, which we can again bound using a (full) variational distribution and the Jensen's inequality:
\begin{equation}
\log \sum_{\bi \in W_{\bk}} \pi_{\bi,z}=\log \sum_{\bi \in W_{\bk}} r_{\bi,\bk,z}^t\frac{\pi_{\bi,z}}{r_{\bi,\bk,z}^t} \ge \sum_{\bi \in W_{\bk}} r_{\bi,\bk,z}^t \log \frac{\pi_{\bi,z}}{r_{\bi,\bk,z}^t}, \label{eq:rBound}
\end{equation}
where $r_{\bi,\bk,z}^t$ is a distribution over locations $\bi$, i.e. $r$ is positive and $\sum_{\bi \in W_{\bk}} r_{\bi,\bk,z}^t = 1$. It is indexed by $\bk$ as the normalization is done differently in each window, it is indexed by $z$ as it can be different for different features, and it is indexed by $t$ as the term is inside the summation over $t$, so a different distribution $r$ could be needed for different bags $\{c^t_z\}$. This distribution could be thought of as information about what proportion of the $c_z$ features of type $z$ was contributed by each of the different sources $\pi_{\bi,z}$ in the window $W_{\bk}$.
However, by performing constrained optimization (so that $r$ adds up to one), we find that assuming a fixed set of parameters $\pi$, the distribution $r_{\bi,\bk,z}^t$ that maximizes the bound is the same for each bag:
\begin{equation}
r_{\bi,\bk,z}^t=\frac{\pi_{\bi,z}}{\sum_{\bi \in W_{\bk}} \pi_{\bi,z}} =\frac{\pi_{\bi,z}}{W_x\cdot W_y \cdot h_{\bk,z}}. \label{eq:r}
\end{equation}
If we do consider distributions $r$ as a feature mapping to the counting grid, then this result is again intuitive. If all we know is that a bag containing $c_z$ features of type $z$ is mapped to the grid section $W_{\bk}$, and have no additional information about what proportions of these $c_z$ features were contributed from different incremental counts $\pi_{\bi,z}$, then the best guess is that these proportions follow the proportions among $\pi_{\bi,z}$ inside the window.

If we assume now that $r$ and $q$ distributions are fixed, then combining Eq. \ref{eq:bound} and Eq. \ref{eq:rBound} and minimizing the resulting bound wrt parameters $\pi_{\bi,z}$ under the normalization constraint over features $z$, we obtain the update rule
\begin{equation}
\hat{\pi}_{\bi,z} \propto \sum_{t=1}^T \sum_{\bk | \bi \in W_\bk} q(\ell^t = \bk) \cdot c_z^t \cdot r_{\bi,\bk,z}^t,
\end{equation}
which by Eq. \ref{eq:r} reduces to
\begin{equation}
\hat{\pi}_{\bi,z} \propto \pi_{\bi,z}^{old} \cdot \sum_{t=1}^T \Big( c_z^t\cdot \sum_{\bk | \bi \in W_\bk} \frac{q(\ell^t = \bk)}{h_{\bk,z}} \;\Big), \label{eq:M1}
\end{equation}
where $ \pi_{\bi,z}^{old} $ is the counting grid at the previous iteration.

The reader will note that in the above, we simply optimized the likelihood of the set of data for a single set of weights $\pi_\bi,z$, as Eq. \ref{eq:bound} is the variational bound for the model with a fixed $\pi$ as it was expanded in the previous section. Thus the iteration of the above equations would optimize for a set of parameters $\pi$ given the observed data and ignoring the prior over $\pi$ in the full network in Fig.  \ref{fig:gm}-\emph{i)}. Of course, the Dirichlet prior with parameters $\eta$ is the appropriate conjugate prior (as in LDA models) making the inclusion of its influence trivial: The parameters $\eta_z$, one for each feature, act as pseudocounts of each feature, 
\begin{equation}
\hat{\pi}_{\bi,z} \propto \eta_z+\pi_{\bi,z}^{old} \cdot \sum_{t=1}^T \Big( c_z^t\cdot \sum_{\bk | \bi \in W_\bk} \frac{q(\ell^t = \bk)}{h_{\bk,z}} \;\Big), \label{eq:M1_full}
\end{equation}
The prior elegantly precludes zero counts of any feature anywhere in $\pi$, preventing overtraining and numerical problems. The innermost sum of the equation above is carried out across all the locations $\bk$ whose window $W_\bk$ contains the generic location $\bi$ indexed in the LHS. This simply reduces to summing in ``shifted'' windows where now $\bk$ represents the lower right corner. 
Finally, by taking derivatives with respect to prior probabilities of different locations, we can readily show that the update for the prior over locations should be updated as follows:
\begin{equation}
P(\ell = \bk)  \propto  \sum_{t=1}^T q(\ell^t = \bk ) .\label{eq:pl1} 
\end{equation}
This is not surprising, as mathematically, the model is a mixture of distributions $h$ and the above is essentially an update for a mixture prior. 
However, if we consider the data efficiency of this update, we see that it differs dramatically from how efficiently the data is used to learn distributions $h$. Consider a large CG model, e.g. on a $64\times 64$ grid, that uses relatively large windows, too, e.g. $16\times 16$. Then even though there are over $4k$ individual distributions $\pi$ to learn, these are in fact used in aggregates of $256$ at a time in each of the $h$ distributions, which makes the parameters of the mixture's sources highly tied. In fact, we can only tile $4\times4=16$ non-overlapping windows over the grid, and the rest of the 4096 overlapping windows are a special kind of interpolation of these $16$. Thus the equivalent capacity of such a model, when compared with a simple mixture, is only 16, allowing such grids to be trained without overtraining with just an order or two more data than this capacity number. In other words, we should be able to train a model with $4k$ fractional sources $\pi$  with only around $1k$ bags of words.  But the equivalently efficient use of data for estimating which parts of the grid are used more than others would require a similar aggregation of fractional probabilities of individual cells, just like $\pi$ distributions are aggregated into $h$ distributions.The similar issue was resolved in epitome models by literally aggregating the updates above within overlapping windows, to avoid overfocusing the prior probability over the $4k$ windows in our example on only those $1k$ positions where training data fell:
\begin{equation}
P( \ell = \bk )  \propto  \sum_{t=1}^T \sum_{\bi \in \bE} q(\ell^t=\bi)\cdot m_{\bk-\bi}. \label{eq:pl2}
\end{equation}
where $m$ is a $E_x \times E_y$ mask, with ones in the upper left corner's $W_x \times W_y$ entries and zeros elsewhere. In our experiments the same update proved to be a valid way to overcome local minima when the prior over location is learned.\footnote{It is also possible to change the model in  way that would allow for this update to arise naturally, in a manner equivalent to defining $h$ distributions as arising from $\pi$ distributions} In this update, the prior $P(\ell)$ must, of course, be normalized across the locations.

The steps in Eqs. \ref{eq:qkl}, \ref{eq:M1} and \ref{eq:pl2} constitute the E and M step which can be iterated till convergence (within a desired precision $\tau$). The learning algorithm is summarized in Alg. \ref{DB}.
\begin{algorithm}
\KwIn{Bag of features, $c_z^t$ for each patch, counting grid size $\bE$, window size $\bW$}
\While{Convergence}{
\texttt{ \% E-Step} \;
\ForEach{Sample $t=1\dots T$}{
$1.\;$ Update $q(\ell^t = \bk )\propto \exp \big\{ \; \sum_z c^t_z \log h_{\bk,z} \; \big\},$\;}
\texttt{ \% M-Step} \;
$2.\;$ Update $\pi_{\bi,z} \propto \pi_{\bi,z}^{old} \cdot \sum_t c_z^t \sum_{\bk | \bi \in W_{\bk}} \frac{q(\ell^t = \bk )}{h_{\bk,z}}$ \;
$3.\;$ Compute $h_{\bk,z} = \frac{1}{W_x\cdot W_y} \sum_{\bi \in W_{\bk}} \pi_{\bi,z}$\;
$4.\;$ Update $P( \ell )$ using Eq. \ref{eq:pl1} or Eq. \ref{eq:pl2}\;
$5.\;$ Compute the Log-Likelihood B with Eq. \ref{eq:bound} \;
$6.\;$ Check for convergence, e.g. $|B-B^{old}|\leq \tau$ \;
}
$6.\;$ Return $\pi_{\bi,z}$, $P(\ell)$ and $\left\{ q(\ell^t) \right\}_t$ \;
\caption{EM-Algorithm to learn a counting grid. \label{DB}}
\end{algorithm}

Starting with non-informative (but symmetry breaking) initialization, this iterative process will jointly estimate the counting grid and align all bags to it. To avoid severe local minima, it is important, however, to consider the counting grid as a torus, and consider all windowing operations accordingly, as was previously proposed for learning epitomes \cite{epitome,epitomeLr,epitomeSca}. This prevents the problems with grid boundaries which otherwise not be crossed when more space is need to grow the layout of the features.

\section{From counting grids to feature epitomes}
\begin{figure}[t]
\centering
\includegraphics[width=1\columnwidth]{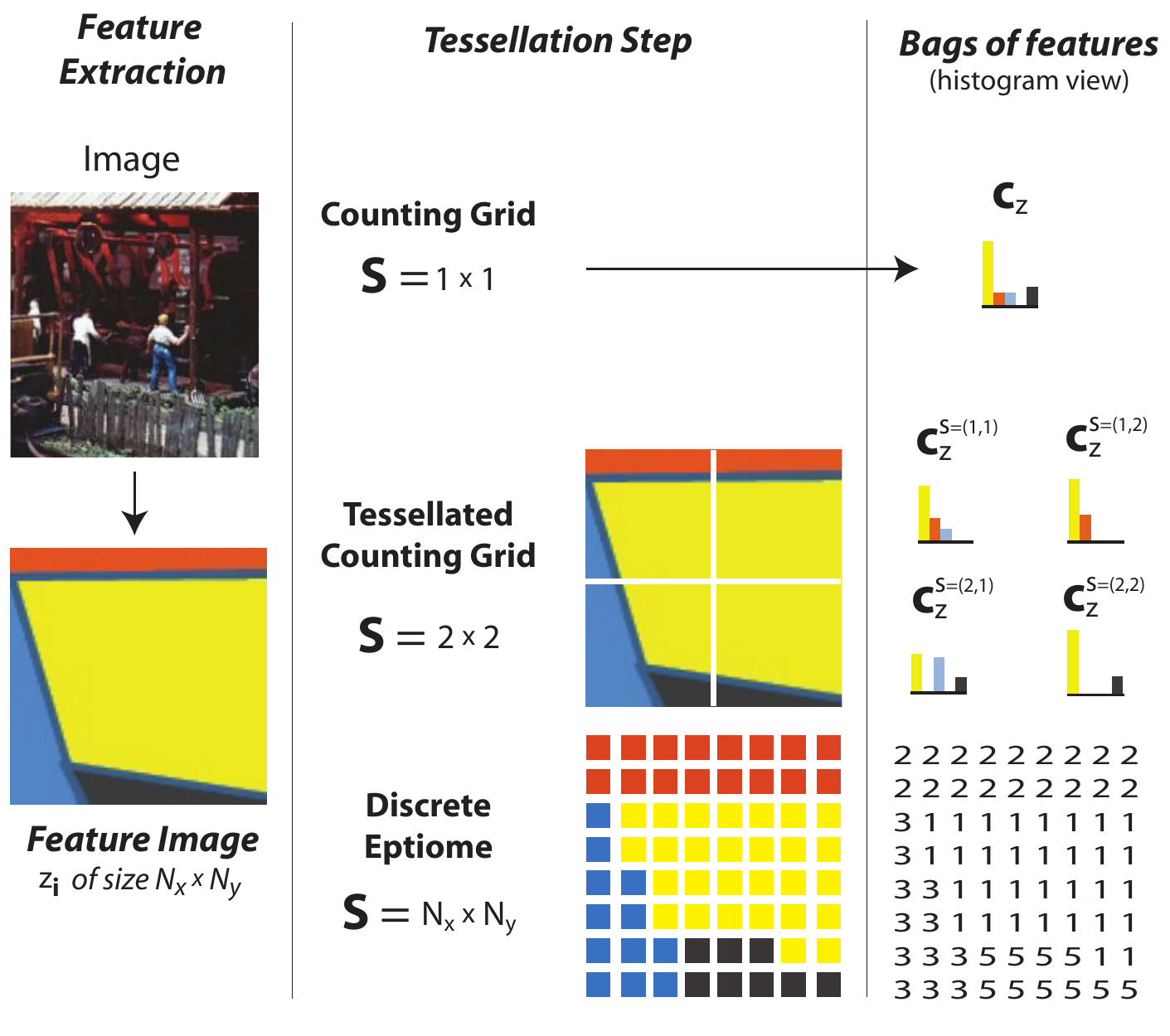}
\caption{Tessellation step illustration. Once the features are extracted and quantized one can decide tessellation $\bS = S_x \times S_y$ of the image and compute the feature counts separately in each different section as illustrated by the second and third column. To the limit, when $\bS = \bW = N_x \times N_y$, we obtain the discrete feature epitome model.}
\label{fig:tess}
\end{figure}

\begin{table}
\label{tab:tab1}
\caption{Relationship between counting grids and other computer vision methods}
\begin{tabular}[h!]{l|c|c|c|c}
Name & E-Step & M-Step & \bW & \bS \\
\hline
Counting grid  & Eq.\ref{eq:qkl} &  Eq.\ref{eq:M1} & $\geq 2\times2$ & $1\times1$ \\
Tessellated CG & Eq.\ref{eq:qkl2} &  Eq.\ref{eq:M2} & $\geq 2\times2$ & $\geq 2\times2$ \\
CG-Epitome \cite{cgCvpr} & Eq.\ref{eq:qkl} &  Eq.\ref{eq:ME} & $N_x\times Ny$ & $\geq1\times1$ \\
Discr. Epit. \cite{epitomeLr} & Eq.\ref{eq:qklE} &  Eq.\ref{eq:ME} & $N_x\times Ny$ & $N_x\times Ny$  \\
Mix.Unigram \cite{mixtureUnigrams} & Eq.\ref{eq:qkl} &  Eq.\ref{eq:M1} & $1\times 1$ & $1\times1$ \\
Spatial BoW \cite{bowReview} & Eq.\ref{eq:qkl} &  Eq.\ref{eq:M1} & $1\times 1$ & $\geq 2\times2$
\end{tabular}
\end{table}
We can express many other models used in vision as special cases of our framework by assuming an appropriate choice of the tessellation $\bS$ of the images and the window size $\bW$. A tessellation $\bS$ is simply a partition of the image space as illustrated in Fig. \ref{fig:tess}. \\

\paragraph*{\textbf{Tessellated counting grids}}
Algorithm \ref{DB} works remarkably well given that its task is essentially to infer a image, not from many image patches as is the case for epitome models, but only from the bag of features representation of for such patches. The task is formidable because no directionality is provided in the bag representation. Unfortunately the iterative algorithm may start to lay out the features topologically correctly, but following inconsistent directions in different parts of the counting grid, leading to local minima (This will be illustrated in the next section). \\
However, we can modify the model and its E and M rules to deal with image representations that consist not of one, but \emph{several} bags of words, each corresponding to a section of the image. In this case feature re-arrangement is tolerated within each region, but the regions themselves cannot move relatively to each other and the model becomes similar in spirit to \cite{bowReview,bowRec}
\\

More specifically, we define a tessellation $\bS = S_x \times S_y$ and for each feature map $z^t_{\bi}$, we compute the feature counts separately in each different section $\{c^{t,\bs}_z\}$ being $\bs = s_x \times s_y$  a bi-dimensional index that runs across the sectors of $\bS$. This process is illustrated in Fig. \ref{fig:tess}-\emph{ii)}. When inferring the mapping of the set of section bags, the window $W_{\bk}$ is tessellated into $S_x \times S_y$ sections of size $W^\bS$ indexed by $W_{\bk}^{\bs}$ int the same way images are tessellated. The histogram comparisons are done accordingly, in formulae:
\begin{equation}
q( \ell^t = \bk )  \propto P(\ell= \bk)\cdot \exp \Big( \sum_{\bs \in \bS} \sum_{z=1}^Z c^{t,\bs}_z \log \sum_{\bi \in W_{\bk}^{\bs}} \pi_{\bi,z} \Big), \label{eq:qkl2}
\end{equation}
It is important to note that all the $S_x \cdot S_y$ bags contribute to the same mapping on the grid. Therefore the tessellated counting grid model inherits the same componential nature of the counting grid while making use of more spatial information, as reported in Tab. \ref{tab:tabintro}. \\
The M step using section bags reduces to
\begin{eqnarray} \label{eq:M2}
\pi_{\bi,z} \propto \pi_{\bi,z}^{old}\cdot  \sum_{t=1}^T \Big( \sum_{\bs \in \bS} c_z^{t,\bs}\cdot \sum_{\bk | \bi \in W^\bS_\bk} \frac{q( \ell^t = \bk )}{h^\bs_{\bk,z}} \;\Big) \label{eq:M2}
\end{eqnarray}
The three plates in Fig. \ref{fig:figure2} show that even just considering an representation consisting of four bags of features for the 4 image sections (upper left, upper right, lower left and lower right) provides enough symmetry breaking that good counting grids can be estimated. \\

\paragraph*{\textbf{Discrete Epitomes}} To the limit, when both tessellation and window size are equal to the images size, e.g., $\bS = \bW = N_x \times N_y$, we obtain the discrete feature epitome model. In this case, each bag is composed by a single feature $c_z^{t,\bs} = z_\bs^t$ and the sector index $\bs$ indexes a pixel $\bi$. In the M-step, there is no re-arrangement of the features in the window and they are simply ``copied'' according to the mappings $q(\ell^t)$. \\
The E-Step thus becomes:
\begin{equation}
q( \ell^t = \bk ) \propto P(\ell= \bk)\cdot \exp \Big( \sum_{\bi \in \bE} \sum_{z=1}^Z [z_\bi^t = z]\cdot \log \pi_{\bk-\bi,z} \Big) \label{eq:qklE}
\end{equation}
where $[\cdot]$ is the indicator function, equal to 1 when the equality holds, zero otherwise. Eq. \ref{eq:qklE} can be efficiently computed using FFTs \cite{tmg}. \\
The M-Step reduces to
\begin{equation}
\pi_{\bi,z} \propto \sum_{t=1}^T \sum_{\bk \in \bE} q( \ell^t = \bk ) \cdot[ z^t_{\bi - \bk} = z] \label{eq:ME},
\end{equation}
Likewise the epitome \cite{epitome} and the counting grid, discrete eptiomes are single-component models. However differently from the former, they are characterized by a multinomial observation model and differently from the latter, they consider the original feature layout making the model less efficient and harder to generalize.
 
Finally, in \cite{epitomeLr} local histograms are used as pixel descriptor. This helped to overcome the rigidity of epitome models and reach great performances on location recognition. This technique loosely correspond to $\bW = N_x \times N_y > \bS$. \\
% A: Controlla!
\begin{figure*}[t]
\centering
\includegraphics[width=0.9\textwidth]{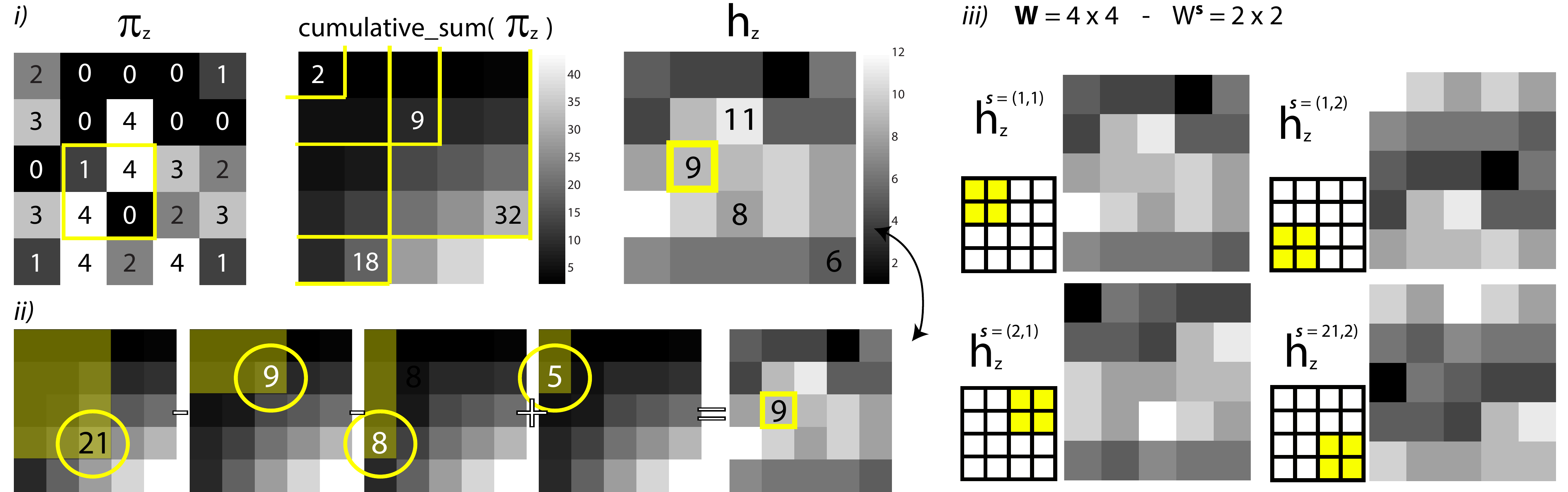}
\caption{Few implementation details useful for computational efficiency. Panles i) and ii) shows how $h$ can be efficiently computed using cumulative sums of $\pi$. Panel iii) show the shifted versions of $h$. }
\label{fig:efficiency}
\end{figure*}

\paragraph*{\textbf{Hybrid counting grid - epitome}} Another alternative is to use the layout of features $z^t_{\bi}$ of each image when updating the counting grid (Eq.\ref{eq:ME}) while its bag of words representation to compute the mapping (Eq.\ref{eq:qkl}). The result is an hybrid between counting grids and epitomes and it is what has been used in the experimental section of the conference version of this paper \cite{cgCvpr}. For some dataset this strategy proved to be successful.

\paragraph*{\textbf{Relationships with other models}} When $\bW = 1\times 1$, the model collapses into a mixture of unigrams \cite{bow} and each point in the grid $\pi_{\bk,z} = h_{\bk,z}$ is now a mixture component. If a tessellation is also enforced, the model becomes similar to the spatial BOW models introduced in \cite{bowRec,bowReview}. \\
Finally, despite the counting grid shares its focus on modeling image feature counts with LDA (and in general topic models), neither model is a generalization of another. However, by using large windows to collate many grid distributions from a large grid, the counting grid model can be thought as a very large mixture of sources without overtraining, as these sources
are highly correlated: Small shifts in the grid change the window distribution only slightly. LDA model does not have this benefit, and thus has to deal with a smaller number of topics to avoid overtraining. Topic mixing cannot quite
appropriately represent feature correlations due to traslational camera motion.

The relationships between counting grids variations introduced in this section in terms of $\bW$, $\bS$ and variational updates are summarized in Tab. \ref{tab:tab1}. 

\section{Computational Complexity and implementation}

Careful examination of the steps reveals that by the efficient use of cumulative sums, all versions of the E and M steps has $\mathcal{O}(N)$ complexity in the size of the counting grid, except for the epitome version. This last version of the counting grid update utilizes the feature layout of the original images $z^t_{\bi}$, which requires the a convolution operation, still manageable in a $\mathcal{O}(N\log N)$ complexity. \\

More generally most of the updates of the E and M steps of the algorithm require computing windowed sums
\begin{equation}
\sum_{(i_x,i_y) \in W_{(k_x,k_y)}}  \hspace{-0.3cm} \pi_{(i_x,i_y),z}
\end{equation} 
where in the previous formula we explicated the two coordinates of the generic position indeces $\bk = (k_x,k_y)$ and $\bi = (i_x,i_y)$. In Fig.\ref{fig:efficiency}-\emph{i)} we show a ``slice'' of $\pi$ and we want to compute the sum in the yellow window. 
These sums can be done efficiently by first computing, in linear time, the cumulative sum 
\begin{equation}
\text{\texttt{cumulative}\textunderscore \texttt{sum}}( \pi_{(k_x,k_y)} ) =\hspace{-0.3cm}\sum_{(i_x,i_y) \le (k_x,k_y)} \hspace{-0.3cm} \pi_{(i_x,i_y)} 
\end{equation}
as illustrated in the second panel of Fig.\ref{fig:efficiency}-\emph{i)}, and then setting 
\begin{eqnarray}
\sum_{(i_x,i_y) \in W_{(k_x,k_y)}}  \hspace{-0.3cm} f_{(i_x,i_y)} =& &  \hspace{-0.5cm} F_{(k_x+W_x+1,k_y+W_y+1)} \nonumber \\
& - & F_{(k_x,k_y+W_y+1)} \nonumber \\
& - & F_{(k_x+W_x+1,k_y)} \nonumber \\
& +&  F_{(k_x,k_y)} 
\end{eqnarray}
which is illustrated by Fig. \ref{fig:efficiency}-\emph{ii)}. 
This procedure is used to compute all window histograms $h$ in the counting grid, as well as in either of the M step versions Eqs. \ref{eq:M1} and \ref{eq:M2}, which only use the counts $c^{t}_z$, and not the original feature layout $z^t_{i,j}$.\\

Efficiency of the computation over multiple section bags in Eqs. \ref{eq:qkl2}, \ref{eq:M2} can be increased if the sections break the window uniformly along both directions. In this case,  one can pre-compute the sum $\sum_{\bi \in W_{\bk}^{\bs}} \pi_{\bi,z}$ in each section and keep $h^s_{k_x+\tau_s,k_y+\tau_s,z}$'s which are shifted versions of each other as Fig. \ref{fig:efficiency}-\emph{iii)} (remember that each sector contribute to the same mapping!).

\begin{figure*}[t]
\centering
\includegraphics[width=0.9\textwidth]{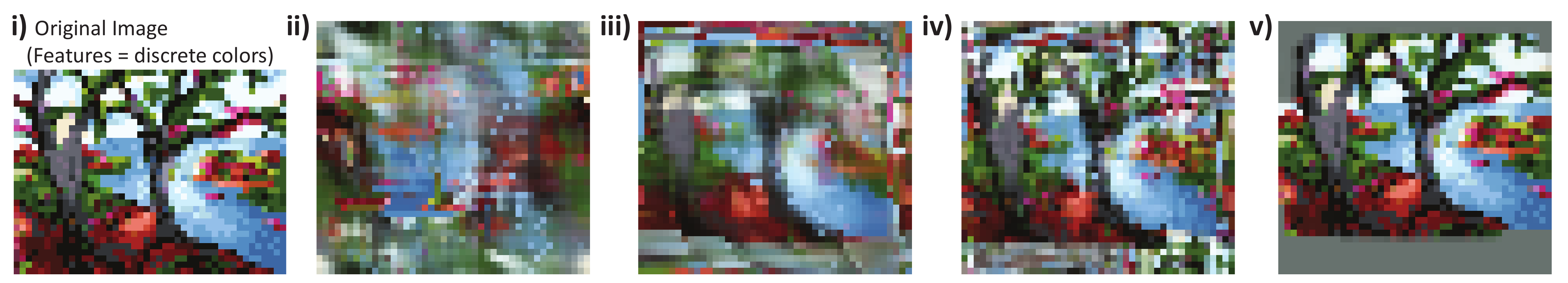}
\caption{The source of 50 image patches taken from random locations i), and counting grids estimated by various versions of the algorithm. Most remarkably ii) is the reconstruction obtained using \emph{only} 50 histograms of image features, and for reconstruction in iii) we used only  50 sets of 4 histograms (from $2 \times 2$ sections of the input images). iv) Result using 16 histograms, (from  $4 \times 4$ sections of the input images) v) Result using cg-epitomes. In all the cases, colors were treated as unrelated 64 discrete features. SEE THE VIDEOS IN THE ADDITIONAL MATERIAL! }
\label{fig:ill2}
\end{figure*}
\section{Layout Reconstruction}\footnote{\textit{In the additional material we added videos that better describe this section and the learning procedure.}}
In scene/object classification tasks, the image features are typically clustered around hundreds of centers and image locations $\bi$ are associated with pointers $z$ to these discretized features. For example, in our classification experiments below, we clustered SIFT \cite{sift} features in Z=200 visual words. The illustrations in Fig. \ref{fig:figure1} and Fig. \ref{fig:figure2} do not provide enough insight into how well the counting grids can be inferred when such large sets of features are considered. Visualizing the feature identities on a grid is difficult, and so, in order to simply study the properties of the counting grid estimation procedures discussed above, we have run the first set of tests on fifty $16 \times 16$ color patches taken at random from a drawing (available in Matlab: load trees) sub-sampled to the resolution of $33 \times 40$. The drawing is illustrated in Fig. \ref{fig:ill2}-\emph{i)}. \\
The patches are first transformed into feature maps $z^t_{\bi}$ pointing to one of Z=64 colors obtained by approximating the color map. Then, $1\times 1$, $2\times 2$ and $4\times 4$ histograms were computed in the appropriate sections of these images to obtain the section bags of words for the algorithm defined by the appropriate equations (see Tab. \ref{tab:tab1}). The algorithm is then run on each section bag representation separately, to obtain the counting grids in \emph{ii)}, \emph{iii)}, and \emph{iv)}. Finally, the plate \emph{v)} shows the result of the combination of the counting grid E step, i.e. mapping of the windows based only on the single bag of words, Eq. \ref{eq:qkl}, and the epitome M step, Eq. \ref{eq:ME}, which uses the known layout of features $z_{\bi}$ in counting grid re-estimate under the assumption that this layout could help arrange features in the counting grid even more than a coarse tessellation. \\

To visualize the different counting grids, each counting grid location $\bk$ was assigned the color equal to the average of the Z=64 colors in color map, weighted by the normalized local feature counts $\pi_{\bk,z}$. The image in \emph{ii)} is therefore an attempt at reconstructing the image in \emph{i)} from fifty color histograms for which we did not provide any additional information about their source: Image \emph{i)} was not provided to the algorithm, nor were the locations of the images from which the fifty histograms were extracted. Note also that the algorithm is not aware of any similarities among the 64 colors, as these are treated as discrete features. \\
Remarkably, a lot of the spatial structure in feature distributions was reconstructed from these 50 histograms. The algorithm discovers that the dark, red and brown tones go together and that they are bordered by green. Elongated dark structures against the blue background are discovered, as is the coast/island boundary. In this sense, the counting grid provides a good model for interpolating among the original 50 histograms, as the histograms from the original image are also likely under the inferred counting grid. Using $2\times 2$ bags as a representation of images is already sufficient to break some symmetry problems and reconstruct almost the entire scene. This improvement is also remarkable, as in this case, ostensibly very little information about the 50 image patches is used: The source image \emph{i)}, or locations of the 50 patches in it are again {\bf not} available to the algorithm, and the algorithm only uses fifty sets of 4 histograms (upper left, upper right, lower left, lower right) over Z=64 colors found in appropriate sections to reconstruct the island and the trees. The most accurate reconstruction is obtained in \emph{v)} by iterating Eqs. \ref{eq:qkl}, \ref{eq:ME}), which is interesting from the epitome modeling point of view. If the counting grid is considered a feature epitome (as used at low resolutions in \cite{epitomeLr}), from which detailed feature maps $z^t_{\bi}$ are generated, rather than simply bags of features, then the inference step that only considers the patch histograms efficiently replaces the convolutional E step of the epitome model (if it were extended to have feature distribution in each image location, rather than real-valued Gaussian models). Furthermore, in this case we also found that this combination is less prone to local minima than the epitome models or the pure counting grid inference and learning of Eqs. \ref{eq:qkl2}, \ref{eq:M2}. Finally we note here that in the extreme case of tessellating the patches down to individual pixels, the counting grid becomes the feature epitome model.

These results are possible, of course due to very high redundancy in images which makes, for example, the extracted $50 \times 64$ count numbers that represent the image patches used for reconstruction of ii) sufficient for this partial recovery of the $33 \times 40 \times \log 64$ parameters necessary to represent i). We next show that these procedures can be used to analyze images that are related by the fact that they belong to the same category, rather than a large image, and that the resulting generalization over the space of possible bag of feature count distributions far surpasses the standard count models including other latent models, such as latent Dirichlet allocation \cite{lda}.

%\begin{figure*}[t]
%\centering
%\includegraphics[width=0.9\textwidth]{corel_images.pdf}
%\caption{Images from the Corel dataset.}
%\label{fig:coreli}
%\end{figure*}

\section{Experimental Section}

In all the experiments as visual words we used SIFT features \cite{sift} clustered into $Z=200$ discretized features. The SIFT processing was based on $16\times 16$ pixel patches spaced 8 pixels apart. In this way, each image was transformed into a feature map $z_{\bi}$ and then its bag of features $c_z$ was created. \\
For a fair comparison, we used our implementations of the reconfigurable part-model \cite{bowRec} and latent Dirichlet allocation \cite{ldaVis} on the very same features. \\
In each task, unless specified, we employed the dataset author's training/testing/validation protocol. To classify a test image we learned a model per class and we assigned the test samples to the class that gives the lowest free energy. \\

We considered counting grids of various complexities with grid size $\bf{E}$ $= [\mathbf{2}\,\, (e.g., 2\times2), \mathbf{3}\,\, (e.g., 3\times3), \dots,\mathbf{10},\mathbf{15},\mathbf{20},\dots,\mathbf{40}]$\footnote{We only considered squared counting grids; where $\bE = \mathbf{N}$ stands for $\bE = N\times N$. The same holds for the window.} and window size $\bf{W} = [2,4,6,\dots]$, limiting the tests only to the combinations with overall capacity $\kappa = \frac{E_x\cdot E_y}{W_x\cdot W_y}$ between 1.5 and $T/2$, where $T$ is the number of training samples. We considered $\bS = [ 1\times 1, 2\times 2, 4\times 4, N_x\times N_y ]$ and we updated $P(\ell)$ with Eq. \ref{eq:pl2}.\\
The capacity $\kappa$ is roughly equivalent to the number of LDA topics as it represents the number of independent windows that they can be fit in the grid; we compared the results using this parallelism \cite{cgCvpr,slcg}. \\

\subsection{Scene Classification}

Scene classification task is useful to shows that counting grids can generalize well even when the most basic spatial interpolation assumption is not perfectly met. In particular we will empirically demonstrate that each individual image can be thought a ``window'' in a larger visual word represented by the counting grid. This has been previously illustrated in Fig. \ref{fig:figure1} where sampling windows gave rise to the features combinations present in the dataset\footnote{With the prior $P(\ell)$ possibly preventing to pick some combination}. \\

\begin{figure}[t]
\centering
\includegraphics[width=0.9\columnwidth]{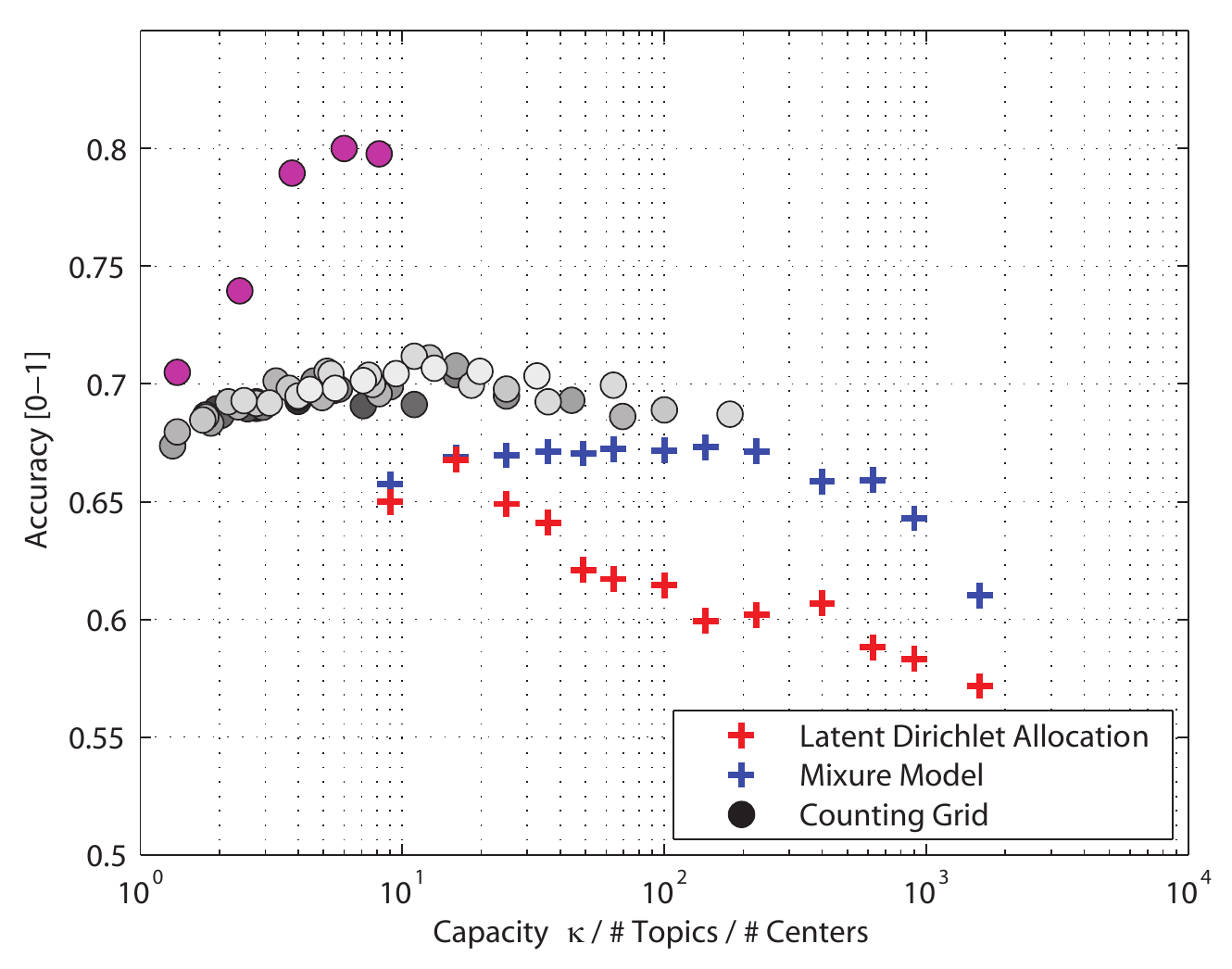}
\caption{15-Scenes classification results. Using pink circles, we also reported the results of \cite{cgCvpr} which are computed using the hybrid CG-Epitome (see Tab.\ref{tab:tab1}). Due to the presence of many training images, the method generalizes very well}
\label{fig:corelr}
\end{figure}

As datasets we considered the 15-Scenes \cite{spk} and the 67-Indoor Scenes \cite{indoor}. Classification accuracy on the former are reported in Fig. \ref{fig:corelr}. On the x-axis we reported the different model complexities, in term of capacity $\kappa$, whereas on the y-axis we reported the accuracy. As the same $\kappa$ can be obtained with different choices of $\bE$ and $\bW$, we specified the counting grid size $\bE$ by using gray levels, the lighter the marker color the bigger the grid. \\
As Fig. \ref{fig:corelr} shows, counting grids performed better than latent Dirichlet allocation. The accuracy regularly increased with $\kappa$, independently from the Grid size $\bE$. It also worth noticing that $P(\ell)$ helped to prevent overtraining for big capacities $\kappa$. In the same figure, we also reported the results of hybrid CG-Epitome approach which comprises the basic CG's E-step and the epitome M-step. For efficiency reasons, we only considered $\kappa = 1.5,2.5,4,6,8$. In this version $\bW = N_x \times N_y$ and the grid size is unequivocally determined by $\kappa$, therefore we used pink markers to show the results. The hybrid CG generalized very well, probably because of the abundance of training data. \\
In Tab. \ref{tab:scenes} we reported a numerical comparison with other models and some discriminative baseline. For counting grids as well as \cite{ldaVis} and \cite{bowRec}, we used 3-Fold crossvalidation on the training set to pick a model complexity.
\begin{table}
 \caption{\textbf{15-Scenes dataset results}}
 \label{tab:scenes}
  \begin{center}
    \vspace{-0.7cm}
\begin{tabular}{lccc}
 \textit{Method} & \textit{Citation} & \textit{Tessellation} & \textit{Accuracy} \\
 \hline
 Mixture Model & & $1\times 1$ & 59,88\% \\
 LDA & \cite{ldaVis} & $1\times 1$ & 65,12\% \\
 Rec. Part Model & \cite{bowRec} & $4\times 4$ & 74,98\% \\
 Spatial BoW & & $4\times 4$ & 73,26\% \\
 \textit{Counting Grid} & & $1\times 1$ & 72,21\%  \\
 \textit{Tess. Counting Grid} & & $4\times 4$ & 74,48\%  \\
 \textit{Hybrid CG-Epitome} & & $N_x\times N_y$ &\textbf{82,79}\%  \\
 \hline
% Linear SVM & & $1\times 1$ & 72,60\% \\
 Spatial Pyramid Kernel & \cite{spk} & $4\times 4$ & 79,93\%  \\
 \end{tabular}
  \end{center}
\end{table}

%\begin{tabular}{ccccc}
%BoW &  LDA  & Rec. Bow. &  CG  & CG-Epitome  \\
%\cite{bow} &  \cite{ldaVis} & \cite{bowRec}  & $\bS = 1\times 1$  & \cite{cgCvpr} \\
%  \hline
%62.10\% & 67,12\% & 76,10\% & 72,21\%  & \textbf{80,41}\%  \\
%\end{tabular}

As second dataset, we considered the 67-indoor scene \cite{indoor} (we did not use the annotations). Results are reported in Tab. \ref{tab:indoor}, where the tessellated counting grid outperformed all the other generative approaches.
\begin{table}
 \caption{\textbf{MIT 67 Indoor Scenes dataset results}}
\label{tab:indoor}
  \begin{center}
  \vspace{-0.7cm}
\begin{tabular}{lccc}
 \textit{Method} & \textit{Citation} & \textit{Tessellation} & \textit{Accuracy} \\
 \hline
 Mixture Model & & $1\times 1$ & 14,31\% \\
 LDA & \cite{ldaVis} & $1\times 1$ & 24,53\% \\
 Rec. Part Model & \cite{bowRec} & $4\times 4$ & 25.32\% \\
 Spatial BoW & & $4\times 4$ & 20,94\% \\
 \textit{Counting Grid} & & $1\times 1$ & 25,42\% \\
 \textit{Tess. Counting Grid} & & $4\times 4$ & \textbf{28,32}\%  \\
 \textit{Hybrid CG-Epitome} & & $N_x\times N_y$ & 16,21\%  \\
  \hline
 % Linear SVM & & $1\times 1$ & 25,12\% \\
 Spatial Pyramid Kernel & \cite{spk} & $4\times 4$ & 32,12\%  \\
 \end{tabular}
  \end{center}
\end{table}

%\begin{table}
% \caption{\textbf{MIT 67 Indoor Scenes dataset Results}. The accuracy of \cite{bowRec} is reported from the original paper}
% \label{tab:indoor}
%  \begin{center}
%\begin{tabular}{ccccc}
%BoW &  LDA  & Rec. Bow. &  CG  & CG  \\
%\cite{bow} &  \cite{ldaVis} & \cite{bowRec}  & $\bS = 1\times 1$  & $\bS = 4\times 4$ \\
%  \hline
%13,82\% & 24,53\% & 27.66\% & 24,42\%  & \textbf{28,32}\%  \\
%\end{tabular}
%  \end{center}
%\end{table}

\subsection{Place Classification}

\begin{figure*}[t!]
\centering
\includegraphics[width=0.9\textwidth]{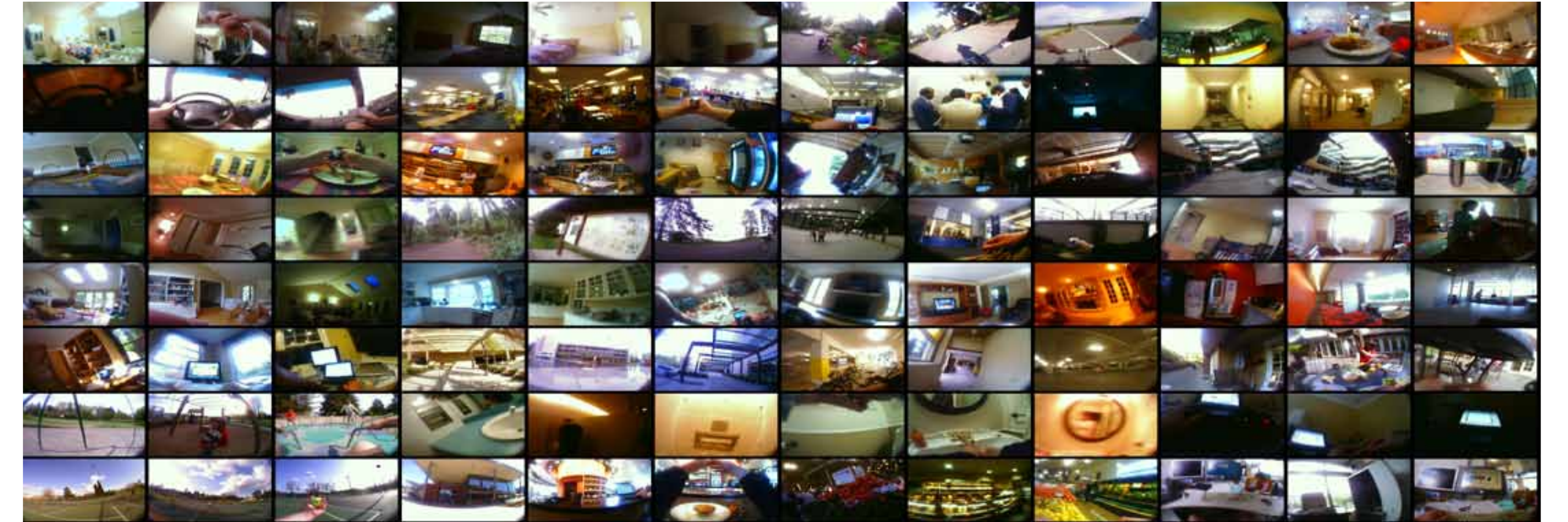}
\caption{Images from the SenseCam dataset.}
\label{fig:sensei}
\end{figure*}

\begin{figure*}[t!]
\centering
\includegraphics[width=1\textwidth]{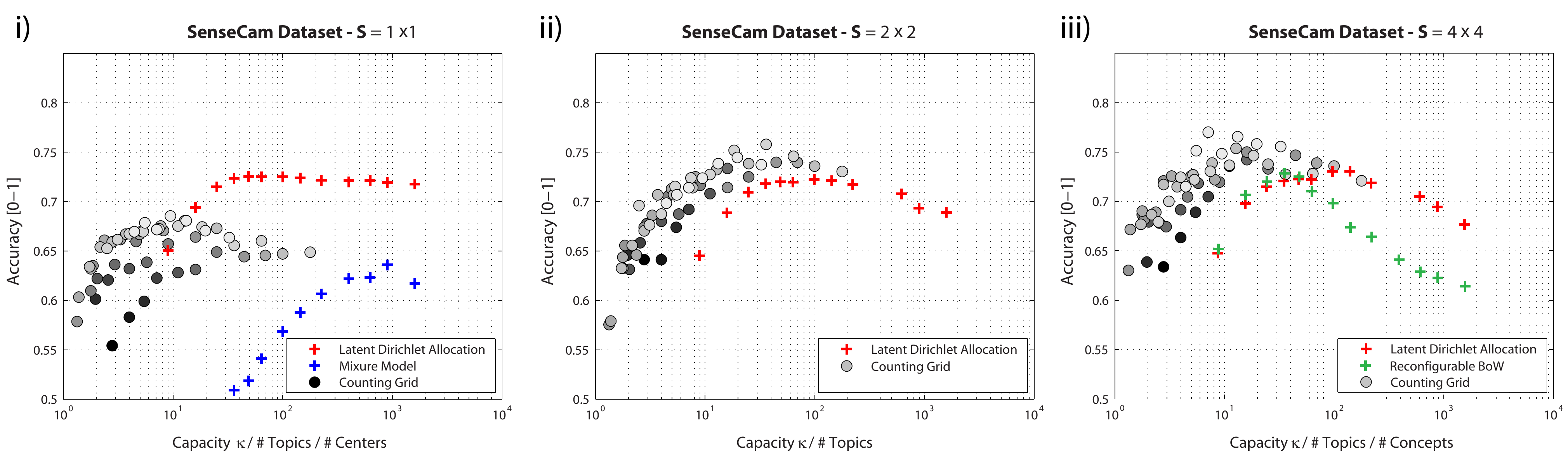}
\caption{Results for SenseCam dataset. i) $\bS = 1 \times 1$. ii) Tessellated version $\bS = 2 \times 2$. iii) Tessellated version $\bS = 4 \times 4$ and comparison with the reconfigurable bag of words model \cite{bowRec} and with latent Dirichlet allocation using the same tessellation.}
\label{fig:senser}
\end{figure*}

Recently in \cite{slcg} a 32-classes dataset have been introduced. This dataset is a subset of the whole visual input of a subject who wore a wearable camera for few weeks. Images in the dataset exhibit dramatic viewing angle, scale, illumination variations and a lot of foreground objects, and clutter. Each category presents images taken in a particular \textit{place} such as house rooms or office environments, or outdoors locations. Some images for each class are shown in Fig. \ref{fig:sensei}.  \\
The task here is place classification. As validation protocol, we used 10-folds cross evaluation. Results are summarized in Fig. \ref{fig:senser}. In the bag-of-word scenario, e.g., $\bS = 1\times 1$,  latent Dirichlet allocation \cite{ldaVis} performed better than regular counting grids and mixture models. This can be explained with local minima issues as some classes have a very limited number of training samples and the counting grid simply cannot well recover the panoramic structure (although this is not perfectly evident or recoverable) of half of the classes. Once we provide some directionality information (coarse tessellations $\bS = 2\times2$)  counting grids can better exploit the panorama and they outperformed significantly LDA \cite{ldaVis} and its naive tessellated extension which learns a model in each sector, summing the $\bS$ likelihoods. Finally in the last panel (Fig. \ref{fig:senser}-\emph{iii)}) we compared $\bS = 4\times4$ tessellated counting grid, again the tessellated latent Dirichlet allocation and the Reconfigurable part model \cite{bowRec} which uses the same spatial information. 
Finer tessellations didn't help recognition but neither hurt up to $\bS = 6\times6$. To the limit, when $\bS = N_x \times N_y$, accuracy does not exceed 30\%. \\

As final experiment on this dataset, we repeated the experiment only using 13 training images per class as previously done in \cite{slcg}. Here we want to test the robustness of the models in overtrain regimes. We reported the final accuracy in Tab.\ref{tab:sense}.\\
\begin{table}
 \caption{\textbf{SenseCam dataset results}}
\label{tab:sense}
  \begin{center}
  \vspace{-0.7cm}
\begin{tabular}{lccc}
 \textit{Method} & \textit{Citation} & \textit{Tessellation} & \textit{Accuracy} \\
 \hline
 Mixture Model  & & $1\times 1$ & 41,19\% \\
 LDA & \cite{ldaVis} & $1\times 1$ & 57,05\% \\
 Rec. Part Model & \cite{bowRec} & $4\times 4$ & 58,17\% \\
 Spatial BoW & & $4\times 4$ & 49,10\% \\
 \textit{Counting Grid} & & $1\times 1$ & 55,32\% \\
 \textit{Tess. Counting Grid} & & $4\times 4$ & \textbf{59,83}\%  \\
 \textit{Hybrid CG-Epitome} & & $N_x\times N_y$ & 39,40\%  \\
  \hline
 % Linear SVM & & $1\times 1$ & 25,12\% \\
 Spatial Pyramid Kernel & \cite{spk} & $4\times 4$ & 52,76\%  \\
 \end{tabular}
 \end{center}
\end{table}

Summarizing counting grids map images onto a bigger real estate, where they lay out the features into a 2D window and stitch overlapping windows trying to recover the panoramic nature of the scene. This fits the qualities of the data acquired by a wearable camera and indeed our model largely outperform \cite{ldaVis,bowRec}. 

\subsection{Wearable Camera Sequences}

\begin{figure}[h!]
\centering
\includegraphics[width=0.9\columnwidth]{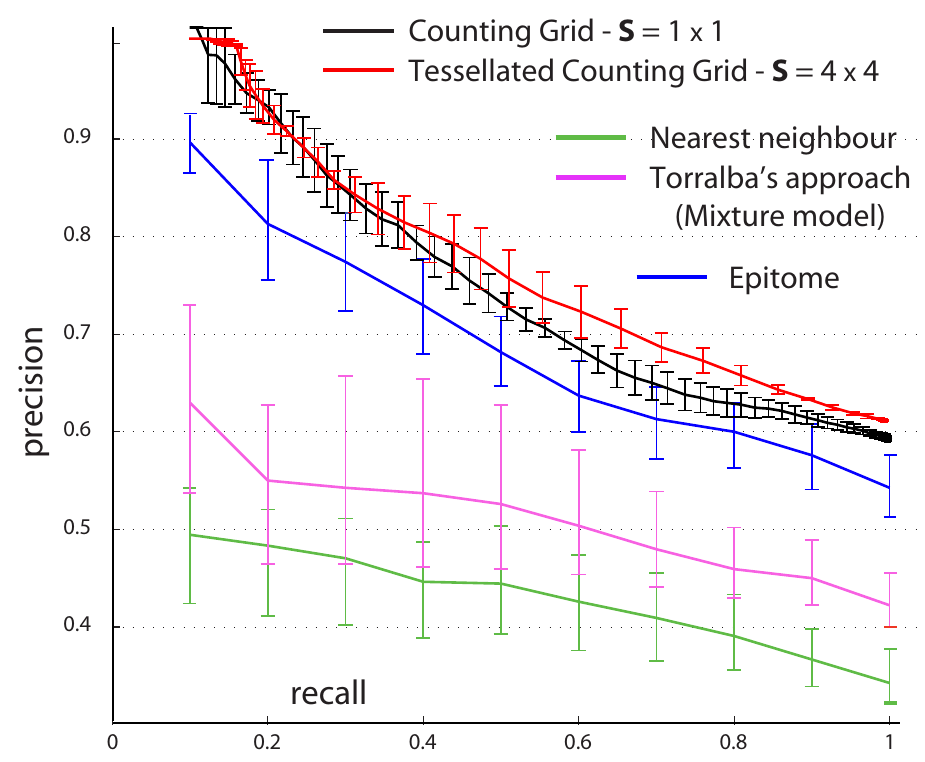}
\caption{Results on Torralba Dataset. We reported the results of \cite{torralbaMoG} (Torralba's approach) and \cite{epitomeLr} (Epitome) from the original papers. We followed the evaluation procedure of \cite{torralbaMoG} and the error bars indicate variability in accuracy across different image sequences.}
\label{fig:torr1}
\end{figure}

We considered the sequences of \cite{torralbaMoG}. This data represents the perfect fit for our model as the true panoramic structure of each scene or place, can actually be recovered. 
The dataset is composed by 7 video sequences acquired with a wearable camera.

The original paper \cite{torralbaMoG} is based on learning a Gaussian mixture model for each class, using Gist \cite{gist} as image descriptor. In addition to \cite{torralbaMoG}, we also compared with Epitomes \cite{epitomeLr} which was, among applications of epitome, one of the most successful. The method of \cite{epitomeLr} uses a low resolution epitome with each low res image location represented by a histogram of features. This method combines several cues: RGB (local) histograms, 
disparity features and Gist. For what concern counting grids, we only used quantized SIFT and we set the complexity of the model using cross-validation considering only models
with capacity $2\leq \kappa \geq 10$. \\

After training a model for each scene $l = 1\dots C$ our goal is to compute the place posterior probabilities for every frame $t$ of the test sequence, given all the previous images $P(l^t = k | \bc^{1:t})$. This can be easily achieved using the forward-backwards procedure \cite{rab}
\begin{eqnarray}
P(l^t = k | \bc^{1:t}) \propto& p( \bc^t | l^t = k)\cdot \sum_j P(l^t=k|l^{t-1}= j) \nonumber \\
& \cdot P(l^{t-1}|\bc^{1:t-1}) \hspace{3cm} \label{eq:hmm}
\end{eqnarray}
We fixed the observation log likelihood to the negative free energy given by our model (Eq. \ref{eq:bound}) while we used EM estimate the transition matrix and the place posteriors. When using HMM, the observation likelihood may be dominated by the transition prior. To balance the contribution we re-scaled the likelihood terms using a constant $\gamma$, chosen via cross-evaluation \cite{deng2003speech}. \\
Results are presented in Fig. \ref{fig:torr1}; the improvement wrt \cite{epitomeLr,torralbaMoG} is significant. The tessellation marginally helped because \emph{i)} training data is abundant and \emph{ii)} the metaphor upon which CGs are based, the ``moving camera'', perfectly fits here. Indeed the spatial layout can be at least piece-wise recovered also from a single bag. \\
Tessellation finer than $\bS = 2\times2$ did not hurt. Latent Dirichlet allocation \cite{lda} and Rec-Bow \cite{bowRec} performances were slightly inferior of \cite{epitomeLr} and we did not report it in the graph for the sake of clarity. \\

\begin{figure*}[t]
\centering
\includegraphics[width=1\textwidth]{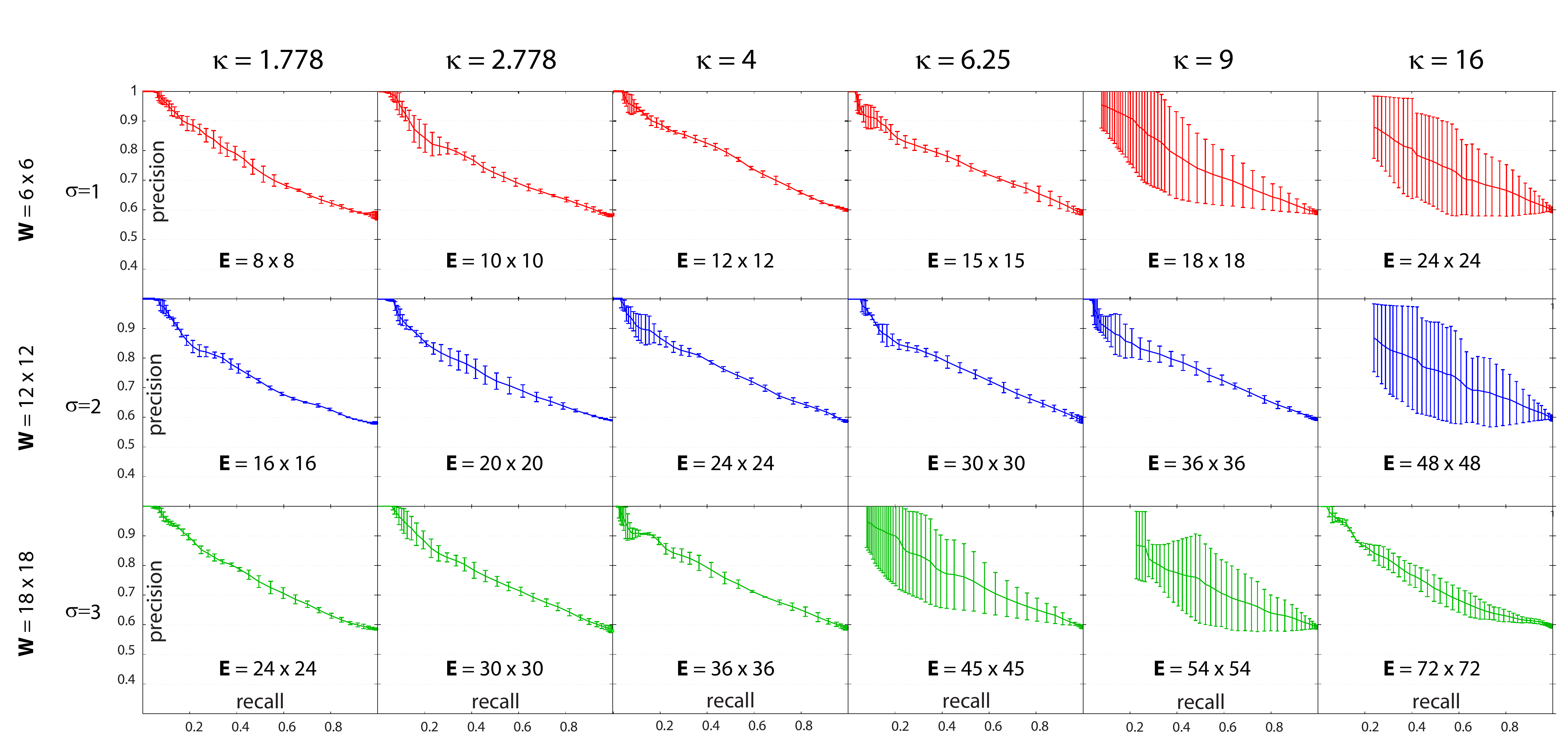}
\caption{Scaled counting grids ($1 \times 1$-case).}
\label{fig:torr2l}
\end{figure*}

We have also investigated what happens if we equally scale $\bE$ and $\bW$. We considered counting grids of size $\bE = \sigma \cdot [\mathbf{8},\mathbf{10},\mathbf{12},\mathbf{15},\mathbf{18},\mathbf{24}]$, $\bW = \sigma\cdot \mathbf{6}$ and three scales $\sigma =1,2,3$ and we run the same experiment on Torralba's sequences. Results are shown in Fig. \ref{fig:torr2l}, where each row represents a different scale.

Results are easily interpretable, counting grids are not very sensitive to the choice of $\bE$ and $\bW$ and what really matters is their ratio $\kappa$. This can also be evinced by Fig. \ref{fig:corelr} and Fig. \ref{fig:senser} where complexities characterized by similar $\kappa$ performed equally well. Higher variances for large $\kappa$, indicate local minima issues. \\
In general, once the window is ``sufficiently big'' for spatial interpolation, scaled models learn ``scaled'' versions of the scene, which are quantitatively (and quantitatively) very similar.  The real estate is too big and the model learn multiple copies of the same scene. \\

We have finally considered a day worth of images from (1800 images ca.) from the SenseCam collection \cite{senseWorkshop} and repeated the same test, combining counting grids an hidden Markov models.
During this day, the camera bearer visited 20 of the 32 labeled locations of the full dataset \cite{slcg}, nevertheless we trained models with all the 32 classes as a-priori we cannot know the locations visited during a day. As for Torralba sequences, our goal is to compute the place posterior probabilities at the instant $t$, given all the previous images, Eq. \ref{eq:hmm} \\
We used at most 30 images per class to learn the models. Results are reported in Tab. \ref{tab:vidd2}. We run \cite{torralbaMoG} using a mixture of dirichlet model over quantized sift histograms (our very same features). For sake of completeness we also implemented the method of \cite{torralbaMoG} extracting the original descriptors from whole images and within the four sectors. In both cases, the performance was below 50\%.

\begin{table}
 \caption{\textbf{Where was I?}}
\label{tab:vidd2}
  \begin{center}
  \vspace{-0.7cm}
\begin{tabular}{lccc}
 \textit{Method} & \textit{Citation} & \textit{Tessellation} & \textit{Accuracy} \\
 \hline
 LDA & \cite{ldaVis} & $1\times 1$ & 76.80\%  \\
 Rec. Part Model & \cite{bowRec} & $4\times 4$ & 74.63\%  \\
 Mixture Model & & $1\times 1$ & 70.37\%  \\
 \textit{Counting Grid} & & $1\times 1$ & 76.21\% \\
 \textit{Tess. Counting Grid} & & $4\times 4$ &  \textbf{81.45}\%  \\
 \end{tabular}
 \end{center}
\end{table}

\subsection{Image clustering on SenseCam}

As final test, we analyzed the same subset of SenseCam, divided in 10 categories used in \cite{epitomeSca}. The images of this subset are suitable for epitomes as they  can actually be stitched together using pixels, therefore a comparison with \cite{epitomeSca,epitome,epitomeLr} is fair. \\
As the spirit of the data collection is to provide summary of the subject's life, we have trained the counting grids in an unsupervised way  (combining images of all categories together) and then investigated if the images are separated in the counting grid in accordance to the human labeling.
We compared with other ``visual summarization'' approaches that lay out the visual input on a larger grid, the epitomic approaches \cite{epitome,epitomeLr,epitomeSca} 
which clusters pixel measurements within an epitome. While in the standard epitomes images are mapped into the epitome by means of pixel wise comparisons, here we are placing bags in a 2 dimensional space, i.e. an image is mapped in a particular spot if its bag-of-word representation agrees with the images mapped in the neighborhood.
To make the comparison fair, we fixed  the complexity of the counting grids to the one used for epitomes in \cite{epitomeSca} (e.g., $\kappa=14$).
Upon learning, each test image is labeled by the label of the closest mapped training image. 
The results are reported in Table \ref{tab:cl}.

\begin{table}
\caption{\textbf{Unsupervised place clustering}}
\label{tab:cl}
  \begin{center}
  \vspace{-0.7cm}
\begin{tabular}{lccc}
 \textit{Method} & \textit{Citation} & \textit{Tessellation} & \textit{Accuracy} \\
 \hline
 Epitome & \cite{epitome} & $N_x \times N_y$ & 69,42\%  \\
 Stel Epitome & \cite{epitomeSca} &  $N_x \times N_y$  & 73,06\%  \\
  LDA & \cite{ldaVis} & $1\times 1$ & 74,32 \% \\
 \textit{Counting Grid} & & $1\times 1$ & 82.34\% \\
 \textit{Tess. Counting Grid} & & $4\times 4$ &  83.94\%  \\
  \textit{Hybrid CG-Epitome} & & $N_x\times N_y$ & \textbf{86,6}\%  \\
   \textit{Feature Epitome} & & $N_x\times N_y$ & 69,93\%  \\
 \end{tabular}
 \end{center}
\end{table}

The counting grid model is so far the best performing model on this task.

\section{Conclusions}

We introduce the counting grid model of images which captures natural constraints on image feature histograms by assuming that these can be represented by averaging of feature distributions from a window into the grid. In this way, the flexibility of the bag of words representation is indirectly enriched by the spatial constraints of epitome-like models. By observing the actual observation model, we see that the counting grid model is not attempting to capture the spatial constraints explicitly as has been often done in the past. In fact, we can view the counting grid as producing a large mixture of histograms whose parameters are constrained in a way that is a natural consequence of the fact that images from which the features are collected live in an ordered 2D space.
Despite their simplicity, both conceptual and algorithmic (the matlab code for counting grid estimation fits half a page), and that the ultimate parametrization used for likelihood computation is simply a set of histograms, this generative model significantly outperforms other histogram-based representations in a variety of tasks and is often approaching the discriminative state of the art (and the features extracted from the generative model can often be used within discriminative models to further improve them \cite{pamiFess}). Computationally, the algorithm is efficient and the computational steps also lend themselves to further improvement of the model to add more scale/rotation reasoning. Experiments show that, despite the apparent need of setting $\bE$ and $\bW$, the algorithm is only sensitive to their ratio. For what concern performances, counting grids, especially in their tessellated version, outperformed standard bag of words approaches in computer vision \cite{ldaVis,bowRec,torralbaMoG,epitomeLr} across most of the datasets considered.
Finally we observe that a variety of methods are based on latent dirichlet allocation and we would like the community considered our method as ``basis'' to solve complex problems or perform complex analysis.

\bibliographystyle{IEEEtran}
\bibliography{AleBIBLIO}

\begin{thebibliography}{10}
\providecommand{\url}[1]{#1}
\csname url@samestyle\endcsname
\providecommand{\newblock}{\relax}
\providecommand{\bibinfo}[2]{#2}
\providecommand{\BIBentrySTDinterwordspacing}{\spaceskip=0pt\relax}
\providecommand{\BIBentryALTinterwordstretchfactor}{4}
\providecommand{\BIBentryALTinterwordspacing}{\spaceskip=\fontdimen2\font plus
\BIBentryALTinterwordstretchfactor\fontdimen3\font minus
  \fontdimen4\font\relax}
\providecommand{\BIBforeignlanguage}[2]{{%
\expandafter\ifx\csname l@#1\endcsname\relax
\typeout{** WARNING: IEEEtran.bst: No hyphenation pattern has been}%
\typeout{** loaded for the language `#1'. Using the pattern for}%
\typeout{** the default language instead.}%
\else
\language=\csname l@#1\endcsname
\fi
#2}}
\providecommand{\BIBdecl}{\relax}
\BIBdecl

\bibitem{bow}
G.~Csurka, C.~R. Dance, L.~Fan, J.~Willamowski, and C.~Bray, ``Visual
  categorization with bags of keypoints,'' in \emph{In Workshop on Statistical
  Learning in Computer Vision, ECCV}, 2004, pp. 1--22.

\bibitem{bowReview}
\BIBentryALTinterwordspacing
J.~Yang, Y.~G. Jiang, A.~G. Hauptmann, and C.~W. Ngo, ``{Evaluating
  bag-of-visual-words representations in scene classification},'' in \emph{MIR
  '07: Proceedings of the international workshop on Workshop on multimedia
  information retrieval}.\hskip 1em plus 0.5em minus 0.4em\relax New York, NY,
  USA: ACM, 2007, pp. 197--206. [Online]. Available:
  \url{http://dx.doi.org/10.1145/1290082.1290111}
\BIBentrySTDinterwordspacing

\bibitem{ldaVis}
L.~Fei-Fei and P.~Perona, ``A bayesian hierarchical model for learning natural
  scene categories.'' in \emph{Proceedings of IEEE Computer Society Conference
  on Computer Vision and Pattern Recognition (CVPR)}, 2005, pp. 524--531.

\bibitem{bowText}
P.~Langley, W.~Iba, and K.~Thompson, ``An analysis of bayesian classifiers,''
  in \emph{Annual Conference on Artificial Intelligence}.\hskip 1em plus 0.5em
  minus 0.4em\relax MIT Press, 1992, pp. 223--228.

\bibitem{5640674}
N.~Bouguila, ``Count data modeling and classification using finite mixtures of
  distributions,'' \emph{Neural Networks, IEEE Transactions on}, vol.~22,
  no.~2, pp. 186 --198, 2011.

\bibitem{mixtureUnigrams}
K.~Nigam, J.~Lafferty, and A.~Mccallum, ``Using maximum entropy for text
  classification,'' in \emph{IJCAI - Workshop on Machine Learning for
  Information Filtering}, 1999.

\bibitem{lda}
D.~Blei, A.~Ng, and M.~Jordan, ``Latent dirichlet allocation,'' \emph{Journal
  of machine Learning Research}, vol.~3, pp. 993--1022, 2003.

\bibitem{pLSA}
T.~Hofmann, ``Probabilistic latent semantic indexing,'' in \emph{Proceedings of
  the annual international ACM conference on Research and development in
  information retrieval (SIGIR)}, 1999, pp. 50--57.

\bibitem{citeulike:2323643}
A.~Bosch, A.~Zisserman, and X.~Munoz, ``Image classification using random
  forests and ferns,'' in \emph{Proceedings of International Conference on
  Computer Vision (ICCV)}, 2007, pp. 1--8.

\bibitem{spk}
S.~Lazebnik, C.~Schmid, and J.~Ponce, ``Beyond bags of features: Spatial
  pyramid matching for recognizing natural scene categories,'' in
  \emph{Proceedings of IEEE Computer Society Conference on Computer Vision and
  Pattern Recognition (CVPR)}, 2006, pp. 2169--2178.

\bibitem{defenseKnn}
O.~Boiman, E.~Shechtman, and M.~Irani, ``In defense of nearest-neighbor based
  image classification,'' in \emph{Proceedings of IEEE Computer Society
  Conference on Computer Vision and Pattern Recognition (CVPR)}, 2008, pp.
  1--8.

\bibitem{vs}
J.~Vogel and B.~Schiele, ``Semantic modeling of natural scenes for
  content-based image retrieval,'' \emph{International Journal of Computer
  Vision}, vol.~72, pp. 133–--157, 2007.

\bibitem{SpatialEnvelope}
A.~Oliva and A.~Torralba, ``Modeling the shape of the scene: A holistic
  representation of the spatial envelope,'' \emph{Int. J. Comput. Vision},
  vol.~42, no.~3, pp. 145--175, 2001.

\bibitem{scPlsa}
A.~Bosch, A.~Zisserman, and X.~Munoz, ``Scene classification via plsa,'' in
  \emph{Proceedings of European Conference on Computer Vision (ECCV)}, 2006,
  pp. 517--530.

\bibitem{4032602}
M.~Boutell, J.~Luo, and C.~Brown, ``Scene parsing using region-based generative
  models,'' \emph{Multimedia, IEEE Transactions on}, vol.~9, no.~1, pp.
  136--146, 2007.

\bibitem{bowRec}
S.~Parizi, J.~Oberlin, and P.~Felzenszwalb, ``Reconfigurable models for scene
  recognition,'' in \emph{Computer Vision and Pattern Recognition (CVPR), 2012
  IEEE Conference on}, 2012, pp. 2775--2782.

\bibitem{torralbaMoG}
A.~Torralba, K.~P. Murphy, W.~T. Freeman, and M.~A. Rubin, ``Context-based
  vision system for place and object recognition,'' in \emph{ICCV}, 2003, pp.
  273--280.

\bibitem{slcg}
A.~Perina and N.~Jojic, ``Spring lattice counting grids: Scene recognition
  using deformable positional constraints.'' in \emph{ECCV (6)}, ser. Lecture
  Notes in Computer Science, A.~W. Fitzgibbon, S.~Lazebnik, P.~Perona, Y.~Sato,
  and C.~Schmid, Eds., vol. 7577.\hskip 1em plus 0.5em minus 0.4em\relax
  Springer, 2012, pp. 837--851.

\bibitem{coates}
\BIBentryALTinterwordspacing
A.~Coates, A.~Y. Ng, and H.~Lee, ``An analysis of single-layer networks in
  unsupervised feature learning,'' in \emph{Proceedings of the Fourteenth
  International Conference on Artificial Intelligence and Statistics, {AISTATS}
  2011, Fort Lauderdale, USA, April 11-13, 2011}, 2011, pp. 215--223. [Online].
  Available:
  \url{http://www.jmlr.org/proceedings/papers/v15/coates11a/coates11a.pdf}
\BIBentrySTDinterwordspacing

\bibitem{labelme}
B.~C. Russell, A.~B. Torralba, K.~P. Murphy, and W.~T. Freeman, ``Labelme: A
  database and web-based tool for image annotation,'' \emph{International
  Journal of Computer Vision}, vol.~77, no. 1-3, pp. 157--173, 2008.

\bibitem{Jojic11}
N.~Jojic and A.~Perina, ``Multidimensional counting grids: Inferring word order
  from disordered bags of words,'' in \emph{{UAI} 2011, Proceedings of the
  Twenty-Seventh Conference on Uncertainty in Artificial Intelligence,
  Barcelona, Spain, July 14-17, 2011}, 2011, pp. 547--556.

\bibitem{cgCvpr}
A.~Perina and N.~Jojic, ``Image analysis by counting on a grid,'' in
  \emph{Proceedings of IEEE Computer Society Conference on Computer Vision and
  Pattern Recognition (CVPR)}, 2011, pp. 1985--1992.

\bibitem{epitomeLr}
K.~Ni, A.~Kannan, A.~Criminisi, and J.~Winn, ``Epitomic location recognition,''
  \emph{IEEE Transactions on Pattern Analysis and Machine Intelligence},
  vol.~31, no.~12, pp. 2158--2167, 2009.

\bibitem{ccg}
A.~Perina and N.~Jojic, ``Capturing layers in image collections with
  componential models: from the layered epitome to the componential counting
  grid,'' in \emph{Proceedings of IEEE Computer Society Conference on Computer
  Vision and Pattern Recognition (CVPR)}, 2013.

\bibitem{DBLP:conf/cvpr/AmerT12}
M.~R. Amer and S.~Todorovic, ``Sum-product networks for modeling activities
  with stochastic structure,'' in \emph{CVPR}, 2012, pp. 1314--1321.

\bibitem{cgFlickr}
P.~Lovato, A.~Perina, N.~Sebe, O.~Zandon{\`a}, A.~Montagnini, M.~Bicego, and
  M.~Cristani, ``Tell me what you like and i'll tell you what you are:
  Discriminating visual preferences on flickr data,'' in \emph{ACCV}, 2012, pp.
  45--56.

\bibitem{epitome}
N.~Jojic, B.~J. Frey, and A.~Kannan, ``Epitomic analysis of appearance and
  shape,'' in \emph{Proceedings of International Conference on Computer Vision
  (ICCV)}, 2003, pp. 34--41.

\bibitem{epitomeSca}
N.~Jojic, A.~Perina, and V.~Murino, ``Structural epitome: a way to summarize
  one's visual experience,'' in \emph{Advances in Neural Information Processing
  Systems}, 2010, pp. 1027--1035.

\bibitem{flex}
N.~Jojic and B.~Frey, ``Learning flexible sprites in video layers,'' in
  \emph{Proceedings of the IEEE Conference on Computer Vision and Pattern
  Recognition}, 2001, pp. 199--206.

\bibitem{gist}
A.~Oliva and A.~Torralba, ``Building the gist of a scene: The role of global
  image features in recognition,'' \emph{Progress in Brain Research: Visual
  perception}, vol. 155, pp. 23--36, 2006.

\bibitem{randomizedSpatialPartition}
Y.~Jiang, J.~Yuan, and G.~Yu, ``Randomized spatial partition for scene
  recognition,'' in \emph{Proceedings of the 12th European conference on
  Computer Vision - Volume Part II}, ser. ECCV'12.\hskip 1em plus 0.5em minus
  0.4em\relax Berlin, Heidelberg: Springer-Verlag, 2012, pp. 730--743.

\bibitem{DBLP:journals/ivc/PerinaCM10}
A.~Perina, M.~Cristani, and V.~Murino, ``Learning natural scene categories by
  selective multi-scale feature extraction,'' \emph{Image Vision Comput.},
  vol.~28, no.~6, pp. 927--939, 2010.

\bibitem{citeulike:9350037}
A.~Perina, N.~Jojic, U.~Castellani, M.~Cristani, and V.~Murino, ``Object
  recognition with hierarchical stel models,'' in \emph{Proceedings of European
  Conference on Computer Vision (ECCV)}, 2010, pp. 15--28.

\bibitem{dpmScenes}
M.~Pandey and S.~Lazebnik, ``Scene recognition and weakly supervised object
  localization with deformable part-based models,'' in \emph{ICCV}, 2011, pp.
  1307--1314.

\bibitem{epitomeSpatialized}
X.~Chu, S.~Yan, L.~Li, K.-L. Chan, and T.~Huang, ``Spatialized epitome and its
  applications,'' in \emph{Computer Vision and Pattern Recognition (CVPR), 2010
  IEEE Conference on}, 2010, pp. 311--318.

\bibitem{prlEpitome}
L.~Bazzani, M.~Cristani, A.~Perina, and V.~Murino, ``Multiple-shot person
  re-identification by chromatic and epitomic analyses,'' \emph{Pattern
  Recognition Letters}, vol.~33, no.~7, pp. 898--903, 2012.

\bibitem{sift}
D.~Lowe, ``Object recognition from local scale-invariant features,'' in
  \emph{Proceedings of International Conference on Computer Vision (ICCV)},
  1999, pp. 1150--1157.

\bibitem{variationalJordan}
M.~Jordan, Z.~Ghahramani, T.~Jaakkola, and L.~Saul, ``An introduction to
  variational methods for graphical models,'' \emph{Machine Learning}, vol.~37,
  no.~2, pp. 183--233, 1999.

\bibitem{variational}
R.~M. Neal and G.~E. Hinton, ``A view of the em algorithm that justifies
  incremental, sparse, and other variants,'' \emph{Learning in graphical
  models}, pp. 355--368, 1999.

\bibitem{tmg}
B.~Frey and N.~Jojic, ``Transformation-invariant clustering using the {EM}
  algorithm,'' \emph{IEEE Transactions on Pattern Analysis and Machine
  Intelligence}, vol.~25, no.~1, pp. 1 -- 17, 2003.

\bibitem{indoor}
A.~Quattoni and A.~Torralba, ``Recognizing indoor scenes,'' in \emph{CVPR},
  2009, pp. 413--420.

\bibitem{rab}
L.~Rabiner, ``A tutorial on {Hidden Markov Models} and selected applications in
  speech recognition,'' \emph{Proc. of IEEE}, vol.~77, no.~2, pp. 257--286,
  1989.

\bibitem{deng2003speech}
\BIBentryALTinterwordspacing
L.~Deng and D.~O'Shaughnessy, \emph{Speech Processing: A Dynamic and
  Optimization-oriented Approach}, ser. Signal Processing and Communications
  Series.\hskip 1em plus 0.5em minus 0.4em\relax Marcel Dekker Incorporated,
  2003. [Online]. Available:
  \url{http://books.google.ca/books?id=136wRmFT\_t8C}
\BIBentrySTDinterwordspacing

\bibitem{senseWorkshop}
\BIBentryALTinterwordspacing
A.~Perina and N.~Jojic, ``In the sight of my wearable camera: Classifying my
  visual experience,'' In 2$^{nd}$ IEEE Workshop on Egocentric Vision, in
  conjunction with CVPR, Tech. Rep., 2012. [Online]. Available:
  \url{http://arxiv.org/abs/1304.7236}
\BIBentrySTDinterwordspacing

\bibitem{pamiFess}
A.~Perina, M.~Cristani, U.~Castellani, V.~Murino, and N.~Jojic, ``Free energy
  score spaces: Using generative information in discriminative classifiers,''
  \emph{IEEE Trans. Pattern Anal. Mach. Intell.}, vol.~34, no.~7, pp.
  1249--1262, 2012.

\end{thebibliography}

% \lstinputlisting{cg_display.m}

\end{document}